\documentclass[twoside,11pt]{article}
\usepackage{epsfig,latexsym,amsbsy,amssymb,amsmath,color, url, ,booktabs, multirow}
\usepackage{graphicx,xspace}
\usepackage{longtable}
\usepackage{mathtools}

\usepackage[colorinlistoftodos,
textwidth=2cm,
textsize=tiny,
linecolor=red!30,
bordercolor=red!30,
backgroundcolor=red!10]{todonotes}

\mathtoolsset{showonlyrefs} 

\allowdisplaybreaks

\usepackage[round]{natbib}
\long\def\acks#1{\vskip 0.3in\noindent{\large\bf Acknowledgments}\vskip 0.2in
	\noindent #1}

\newenvironment{keywords}
{\bgroup\leftskip 20pt\rightskip 20pt \small\noindent{\bf Keywords:} }%
{\par\egroup\vskip 0.25ex}

\graphicspath{{../jmlr/figures/}}

\newcommand{\commentOut}[1]{} 

\newcommand{\ie}{i.e.\xspace}
\newcommand{\eg}{e.g.\xspace}

\newcommand{\iid}{i.i.d.\xspace~}

\newcommand{\mat}[1]{\mathbf{#1}}
\renewcommand{\vec}[1]{ \mathbf{#1} } 
\newcommand{\vecS}[1]{\boldsymbol{ #1 }  } 
\newcommand{\vecentry}[2]{{\mathrm #1_{#2}}}
\newcommand{\matentry}[3]{{\mathrm #1_{#2,#3}}}
\newcommand{\matcol}[2]{\mat{#1}_{\cdot,#2}}

 \newcommand{\A}{\mat{A}}
\newcommand{\B}{\mat{B}}
\newcommand{\C}{\mat{C}}
\newcommand{\D}{\mat{D}}

\newcommand{\I}{\mat{I}}
\newcommand{\K}{\mat{K}}

\newcommand{\X}{\mat{X}}
\newcommand{\W}{\mat{W}}
\newcommand{\Y}{\mat{Y}}
\newcommand{\Z}{\mat{Z}}

\newcommand{\calD}{\mathcal{D}} 
\newcommand{\calL}{\mathcal{L}}

\newcommand{\calO}{\mathcal{O}}

\DeclareMathOperator{\diag}{diag}

\newcommand{\hada}{\odot}

\newcommand{\expectation}[2]{ \mathbb{E}_{#1}{\left[#2\right]} }
\newcommand{\expectationsingle}[1]{ \mathbb{E}_{#1}}
\newcommand{\variance}{\mathbb{V}}
\newcommand{\covariance}{\mathbb{C}\text{ov}}
\newcommand{\Normal}{\mathcal{N}}

\newcommand{\gp}{\mathcal{GP}}
\newcommand{\kernel}{\kappa}

\newcommand{\der}{\text{d}}

\newcommand{\fullderiv}[2]{\frac{d{#1}}{d{#2}}}

\newcommand{\gradient}{\nabla}

\newcommand{\trace}{\mbox{ \rm tr }}
\renewcommand{\det}[1]{\left\lvert#1\right\rvert}
\newcommand{\defeq}{\stackrel{\text{\tiny def}}{=}}

\newcommand{\mth}{\mathrm{th}}

\newcommand{\bigO}{\calO}

\newcommand{\kl}[2]{\mathrm{KL}(#1 \lVert #2)}

\newcommand{\name}[1]{{\textsc{#1}}\xspace}

\newcommand{\evb}{\name{evb}}

\newcommand{\secref}[1]{\S \ref{#1}}
\newcommand{\secrefout}[1]{\S#1}
\newcommand{\apref}[1]{Appendix \ref{#1}}

\newcommand{\gpu}{\name{gpu}}
\newcommand{\cpu}{\name{cpu}}
\newcommand{\gpml}{\name{gpml}}
\newcommand{\gpy}{\name{gpy}}

\newcommand{\mining}{\name{mining}}
\newcommand{\boston}{\name{boston}}
\newcommand{\creep}{\name{creep}}
\newcommand{\abalone}{\name{abalone}}
\newcommand{\cancer}{\name{cancer}}
\newcommand{\usps}{\name{usps}}
\newcommand{\sarcos}{\name{sarcos}} 
\newcommand{\mnist}{\name{mnist}} 
\newcommand{\mnistbin}{\name{mnist-b}}
\newcommand{\mnistlarge}{\name{mnist8m}}
\newcommand{\sarcostwo}{\name{sarcos-2}}

\newcommand{\inla}{\name{inla}}
\newcommand{\bbvi}{\name{bbvi}}
\newcommand{\vi}{\name{vi}}
\newcommand{\mcmc}{\name{mcmc}}
\newcommand{\gptext}{{\sc gp}\xspace}
\newcommand{\savigp}{{\sc savigp}\xspace}

\newcommand{\modg}{{\sc m}o{\sc g}\xspace}
\newcommand{\mog}{{\sc m}o{\sc g}\xspace}
\newcommand{\full}{{\sc fg}\xspace}
\newcommand{\ind}{{\sc ind}\xspace}
\newcommand{\mix}{{\sc m}o{\sc g1}\xspace}
\newcommand{\mixtwo}{{\sc m}o{\sc g2}\xspace}

\newcommand{\sftext}{\name{sf}}
\newcommand{\sfmath}{\text{SF}}
\newcommand{\ard}{\name{ard}}

\newcommand{\wgp}{\name{wgp}}
\newcommand{\gprn}{\name{gprn}}
\newcommand{\lgcp}{\name{lgcp}}
\newcommand{\vbo}{\name{vbo}}
\newcommand{\ep}{\name{ep}}
\newcommand{\vq}{\name{vq}}
\newcommand{\hmc}{\name{hmc}}
\newcommand{\ess}{\name{ess}}
\newcommand{\svigp}{\name{svigp}}
\newcommand{\gpone}{\name{gp1000}}
\newcommand{\gptwo}{\name{gp2000}}
\newcommand{\mc}{\name{mc}}
\newcommand{\bfgs}{\name{bfgs}}
\newcommand{\sgd}{\name{sgd}}
\newcommand{\adadelta}{\name{adadelta}}

\newcommand{\elbotext}{\name{elbo}}
\newcommand{\elbohat}{\widehat{\calL}_{\text{elbo}}}
\newcommand{\elbo}{\calL_{\text{elbo}}}

\newcommand{\ellterm}{\calL_{\text{ell}}}
\newcommand{\klterm}{\calL_{\text{kl}}}
\newcommand{\enterm}{\calL_{\text{ent}}}
\newcommand{\crossterm}{\calL_{\text{cross}}}
\newcommand{\elltermhat}{\widehat{\calL}_{\text{ell}}}  
\newcommand{\entermhat}{\widehat{\calL}_{\text{ent}}}
\newcommand{\elltermhatn}{\elltermhat^{(n)}}
\newcommand{\elltermkn}{\ellterm^{(k,n)}}

\newcommand{\sse}{\name{sse}}
\newcommand{\msse}{\name{msse}}
\newcommand{\nlpd}{\name{nlpd}}
\newcommand{\nlp}{\name{nlp}}
\newcommand{\rmse}{\name{rmse}}
\newcommand{\er}{\name{er}}

\newcommand{\param}{\vecS{\eta}}
\newcommand{\hyperparam}{\vecS{\theta}}
\newcommand{\likeparam}{\vecS{\phi}}
\newcommand{\varparam}{\llambda}
\newcommand{\sfactor}{{\sc \epsilon}} 
\newcommand{\pcond}[2]{p(#1  | #2 )}
\newcommand{\lgpm}{{\sc lgpm}\xspace}

\newcommand{\bs}{\boldsymbol}
\newcommand{\llambda}{\bs{\lambda}}
\newcommand*{\QEDA}{\hfill\ensuremath{\blacksquare}}%

\newcommand{\mydot}{\cdot}

\newcommand{\s}{S} 
\newcommand{\n}{N} 
\newcommand{\m}{M} 
\newcommand{\p}{P} 
\newcommand{\q}{Q} 
\renewcommand{\d}{D} 
\renewcommand{\k}{K} 
\newcommand{\y}{\vec{y}} 

\newcommand{\yn}{\y_{n}} 

\newcommand{\f}{\vec{f}} 
\newcommand{\fn}{\f_{n \mydot}} 
\newcommand{\fj}{\f_{\mydot j}} 
\renewcommand{\u}{\vec{u}} 
\newcommand{\uj}{\vec{u}_{\mydot j} } 
\newcommand{\x}{\vec{x}}
\newcommand{\z}{\vec{z}}
\renewcommand{\X}{\mat{X}}
\newcommand{\Zj}{\mat{Z}_j}
\newcommand{\xstar}{\x_{\star}}
\newcommand{\fstar}{\f_{\star \mydot}}
\newcommand{\fstarj}{f_{\star j}}

\newcommand{\ystar}{\y_{\star}}

\renewcommand{\K}{\mat{K}}
\newcommand{\Kzzall}{\K_{zz}} 
\newcommand{\Kzzallinv}{\K^{-1}_{zz}} 
\newcommand{\Gram}[2]{\K_{#1#2}^j}
\newcommand{\Kzz}{\K_{\z\z}^j} 
\newcommand{\Kxx}{\K_{\x\x}^j}
\newcommand{\Kxxall}{\K_{\x\x}}
\newcommand{\Kxxallinv}{\K_{\x\x}^{-1}}
\newcommand{\Kxxinv}{(\K_{\x\x}^j)^{-1}}
\newcommand{\Kxz}{\K_{\x\z}^j} 
\newcommand{\Kzx}{\K_{\z\x}^j} 
\newcommand{\kzn}{{\vec{k}_{\z n}^j}} 
\newcommand{\kznt}{(\kzn)^T} 
\newcommand{\knz}{\kznt}
\newcommand{\knn}{\kernel(\x_n, \x_n)}
\newcommand{\Kzzinv}{(\K_{\z\z}^j)^{-1}} 
\newcommand{\Aj}{\mat{A}_j}
\newcommand{\priormean}{\tilde{\vecS{\mu}}_j} 
\newcommand{\priorcov}{\widetilde{\K}_j}
\newcommand{\postmean}[1]{\vec{m}_{#1}}
\newcommand{\postcov}[1]{\mat{S}_{#1}}
\newcommand{\postcovdiag}[1]{\tilde{\mat{S}}_{#1}}
\newcommand{\ajn}{\vec{a}_{jn}}
\newcommand{\fnotn}{\f_{\neg n \mydot}}

\newcommand{\ynotn}{\y_{\neg n}}

\newcommand{\predmean}[1]{\mu^{\star}_{ #1}}
\newcommand{\predvar}[1]{\sigma^{\star 2}_{ #1}}

\newcommand{\plike}{\pcond{\y}{\f, \likeparam}}
\newcommand{\logpliken}{\log \pcond{\yn}{\fn, \likeparam}}
\newcommand{\qjoint}{q(\f, \u | \llambda)}
\newcommand{\qu}{q(\u | \llambda)}
\newcommand{\qku}{q_k(\u | \llambda_k)}  
\newcommand{\qf}{q(\f | \llambda)}
\newcommand{\qkf}{q_k(\f | \llambda_k)}  
\newcommand{\qkfstar}{q_k(\fstar | \llambda_k)}  
\newcommand{\qknf}{q_{k(n)}(\fn | \llambda_k )}
\newcommand{\qnf}{q_{(n)}(\fn)}

\newcommand{\qfmean}[1]{\vec{b}_{#1}}        
\newcommand{\qfcov}[1]{\mat{\Sigma}_{#1}}        

\renewcommand{\vecentry}[2]{{[#1]_{#2}}}
\renewcommand{\matentry}[3]{{[#1]_{#2,#3}}}
\newcommand{\matentryinv}[3]{{[#1]^{-1}_{#2,#3}}}
\newcommand{\matentrypow}[4]{{[#1]^{#4}_{#2,#3}}}
\renewcommand{\matcol}[2]{[{#1}]_{:,#2}}

\newcommand{\grad}{\gradient}

\newcommand{\nkl}{\Normal_{k \ell}}
\newcommand{\zk}{z_{k}}
\newcommand{\zl}{z_{\ell}}
\newcommand{\Ckl}{\C_{kl}}
\newcommand{\pil}{\pi_{\ell}}
\newcommand{\tildeCkl}{\tilde{\C}_{kl}}

\newcommand{\qfcovkjn} {\matentry{ \qfcov{k(n)} }{j}{j}}
\newcommand{\qfcovkjninv} {\matentryinv{ \qfcov{k(n)} }{j}{j}}
\newcommand{\qfmeankjn} {\vecentry{ \qfmean{k(n)} }{ j } }
\newcommand{\qfcovkjninvtwo} {\matentrypow{ \qfcov{k(n)} }{j}{j}{-2} }

\newcommand{\fnjki} {f_{nj}^{(k,i)}}

\newcommand{\elln}{\bar{\ell}_n}
\newcommand{\Sigmann}{\matentry{\qfcov{j}}{n}{n}}
\newcommand{\gsj}{\grad_{\postcov{j}}}
\newcommand{\lambdajn}{\lambda_{jn}}
\newcommand{\Lambdaj}{\mat{\Lambda}_j}

\newcommand{\footremember}[2]{%
	\thanks{#2}
	\newcounter{#1}
	\setcounter{#1}{\value{footnote}}%
}
\newcommand{\footrecall}[1]{%
	\footnotemark[\value{#1}]%
}

\title{Generic Inference in Latent Gaussian Process Models}

\author{Edwin V.\ Bonilla\footremember{joint}{Joint first author.}\footremember{data61}{Machine Learning Research Group, Data61, Sydney NSW 2015, Australia.}
     \and
     Karl Krauth\footrecall{joint}
     \footremember{berkeley}{Department of Electrical Engineering and Computer Science, University of California, Berkeley, CA 94720-1776, USA.}
     \and
     Amir Dezfouli\footrecall{data61}
  }

\begin{document}	
	\maketitle
	\maketitle
\begin{abstract}
We develop an automated variational method for inference in  models with Gaussian process (\gptext) priors and general likelihoods. The method supports multiple outputs and multiple latent functions  and does not require detailed knowledge of the conditional likelihood, only needing its evaluation as a black-box function.  Using a mixture of Gaussians as the variational distribution,  we show that the evidence lower bound and its gradients can be estimated efficiently using samples from  \emph{univariate} Gaussian distributions. Furthermore, the method is scalable to large datasets  which is achieved by using an augmented prior via the inducing-variable approach underpinning most sparse \gptext approximations, along with parallel computation and stochastic optimization. 
We evaluate our approach quantitatively and qualitatively with experiments on small datasets, medium-scale datasets and  large datasets, showing its competitiveness under different likelihood models and  sparsity levels.  
%
%
 On the large-scale experiments involving prediction of airline delays and classification of handwritten digits, we show that our method is on par with the state-of-the-art hard-coded approaches for scalable \gptext regression and classification.
 \end{abstract}


\section{Introduction \label{sec:intro}}


Developing fully automated  methods for inference in complex probabilistic models 
has become arguably one of the most exciting areas of research in machine learning, 
with notable examples in the probabilistic programming community
\citep[see e.g.][]{hoffman-gelman-jmlr-2014,wood-et-al-aistats-2014,goodman-et-al-2008,domingos-et-al-aaai-2006}.
Indeed, while these probabilistic programming systems allow  for the formulation of very expressive and flexible 
probabilistic models, building efficient inference methods for reasoning in  such systems is a significant challenge. 

One particularly interesting setting is when the prior over the latent parameters of the model 
can be well described with a Gaussian process  \citep[\gptext,][]{rasmussen-williams-book}.
\gptext priors are a popular  choice in practical non-parametric Bayesian  modeling with 
perhaps the most straightforward and well-known application being the standard 
regression model with Gaussian likelihood, for which the posterior can be computed in closed form
\citep[see e.g.][\secrefout 2.2]{rasmussen-williams-book}. 

Interest in  \gptext models stems  from their functional  non-parametric Bayesian nature and there are at least three critical benefits from adopting such a modeling approach.    
Firstly, by being a prior over functions, they represent a much more elegant way to
address problems such as regression, where it is less natural 
to think of a prior over parameters such as the weights in a linear model.  
Secondly, by being Bayesian, they provide us with a principled way to combine our prior beliefs with 
the observed data in order to predict a full posterior distribution over unknown quantities,  
which of course is much more informative than a single point prediction.
Finally,  by being non-parametric, they address the common concern of 
parametric approaches about how well we can fit the data,  
since the model space is not constrained to have a parametric form.
\subsection{Key Inference Challenges in Gaussian Process Models}
Nevertheless, such a principled and  flexible approach to machine learning comes at 
the cost of facing two fundamental probabilistic inference challenges, 
(i) scalability to large datasets and (ii) dealing with nonlinear likelihoods. 
With regards to the first challenge, scalability, \gptext models have been 
notorious for their poor scalability as a function of the number of training 
points. Despite ground-breaking work in understanding 
scalable \gptext{s} through the so-called sparse 
approximations \citep{quinonero2005unifying} and in developing 
practical inference methods for these models \citep{titsias2009variational},
recent literature on addressing this challenge 
gives  a  clear indication that the scalability problem is still a very active area of research,
see e.g.~\citet{das-et-al-arxiv-2015,deisenroth-ng-icml-2015,hoang-et-al-icml-2015,dong2017scalable,hensman-et-al-jmlr-2017,salimbeni2018natural,john2018large}.

Concerning the second challenge,  dealing with nonlinear likelihoods, 
the main difficulty is that of estimating a posterior over latent functions 
distributed according to a \gptext prior,  given observations assumed to be
 drawn from a possibly nonlinear conditional likelihood model. 
In general, this posterior is analytically intractable and one must resort to 
approximate methods. These methods   can be roughly classified into \emph{stochastic} approaches
 and  \emph{deterministic} approaches. 
The former,  stochastic approaches, are based on sampling algorithms such as 
 Markov Chain Monte Carlo \citep[\mcmc, see e.g.][for an overview]{neal1993probabilistic} and 
 the latter,  deterministic approaches, are based on optimization techniques 
 and include variational inference \citep[\vi,][]{jordan1998introduction}, 
 the Laplace approximation \citep[see e.g.][Chapter 27]{mackay2003information} and expectation propagation \citep[\ep,][]{minka2001expectation}. 
 
 On the one hand, although stochastic techniques such as \mcmc 
 provide a flexible framework for sampling from complex posterior 
 distributions of probabilistic models,
 their generality comes at the expense of a very high computational 
 cost as well as cumbersome convergence analysis. 
On the other hand, deterministic methods such as variational inference 
build upon the main insight that  optimization is generally easier than integration.
Consequently,  
they estimate a posterior by maximizing  a lower bound of the marginal likelihood, 
the so-called evidence lower bound (\elbotext). 
Variational methods  can be considerably faster than \mcmc but  
they  lack \mcmc's broader applicability, usually requiring  
mathematical derivations on a model-by-model basis.%
\footnote{
	However, we note  recent developments such as 
	variational auto-encoders \citep{kingma2013auto} and refer the 
	reader to the related work  and the discussion (Sections 
	\ref{sec:related-recent} and \ref{sec:conclusion}, respectively). 
}
\subsection{Contributions}
In this paper we address the above challenges by developing a scalable  
automated variational method for inference in 
models with Gaussian process priors and general likelihoods. This method   
reduces the overhead of the tedious mathematical 
derivations traditionally inherent to variational algorithms, 
allowing their application to a wide range of problems.
In particular, we consider models 
with multiple latent functions, multiple outputs and 
non-linear likelihoods that satisfy the following properties:
(i)  factorization across latent functions  and 
(ii)  factorization across observations.
The former assumes that, when there are more than one latent function, they are generated from  independent \gptext{s}. 
The latter assumes that, given  the latent functions,  
the observations are  conditionally independent. 

Existing \gptext models, such as regression \citep{rasmussen-williams-book}, 
binary and multi-class classification \citep{nickisch2008approximations,williams1998bayesian}, 
warped \gptext{s} \citep{snelson2003warped}, log Gaussian Cox process \citep{moller1998log}, 
and multi-output regression \citep{wilson-et-al-icml-12}, all fall into  this class of models.
Furthermore, our approach goes well beyond standard settings for 
 which elaborate learning machinery has been developed, 
as we only require access to the likelihood function in a black-box manner.
As we shall see below, our inference method can scale up to very large datasets, 
hence we will refer to it  as \savigp, which stands 
 for \emph{scalable automated variational inference for Gaussian process} models.
The key characteristics of our method  and contributions of this work are  
summarized below. 

\begin{itemize}

\item \emph{Black-box likelihoods:}
As mentioned above, the main contribution of this work  is to be able to 
carry out posterior inference with \gptext priors and general likelihood models, without knowing the details of 
the conditional likelihood (or its gradients) and only requiring its evaluation  as a black-box function.

\item \emph{Scalable inference and stochastic optimization:} 
 By building upon the inducing-variable approach underpinning most
sparse approximations to \gptext models \citep{quinonero2005unifying,titsias2009variational},  
the computational complexity of our method is dominated by
$\bigO(\m^3)$ operations in time, where $\m \ll \n$ is the number of inducing variables and 
$\n$ the number of observations. This 
is in contrast to naive inference in \gptext models which has a time complexity of $\bigO(\n^3)$. 
As the resulting \elbotext decomposes over the training datapoints, 
our model is amenable to parallel computation and stochastic optimization.
In fact,  we provide  
an implementation that can scale up to a very large number of observations, 
exploiting stochastic optimization,  multi-core architectures and \gpu computation.
\item \emph{Joint learning of model parameters:}
As our approach is underpinned by variational inference principles, 
\savigp allows for learning of all model parameters, including 
posterior parameters, inducing inputs, covariance hyperparameters 
and  likelihood parameters, within the same framework via maximization 
of the evidence lower bound (\elbotext). 
\item \emph{Multiple outputs and multiple latent functions:}  \savigp is designed to 
support models with multiple outputs and multiple latent functions, such as
in multi-class classification \citep{williams1998bayesian} and non-stationary multi-output 
regression \citep{wilson-et-al-icml-12}. It does so in a very flexible way, allowing 
the different \gptext priors on the latent functions to have different covariance functions 
and  inducing inputs. 
\item \emph{Flexible posterior:} \savigp uses a mixture of Gaussians as the approximating 
posterior distribution. This is a very general approach as it is well 
known that, with a sufficient number of components, almost any 
continuous density can be approximated with arbitrary accuracy \citep{maz1996approximate}.
\item \emph{Statistical efficiency:} 
By using knowledge of the approximate posterior and the structure 
of the \gptext prior, we exploit the decomposition of the 
\elbotext, into a KL-divergence term and an expected log likelihood term, 
to provide \emph{statistically} efficient parameter estimates. 
In particular,  we derive an analytical lower bound for the KL-divergence term 
and we show  that, for general black-box likelihood models, 
the expected log likelihood term and its gradients can be computed efficiently
using samples from \emph{univariate} Gaussian distributions.
\item \emph{Efficient re-parametrization:} 
For the case of a single full Gaussian variational posterior, 
we show that it is possible to re-parametrize the model 
so that the optimal posterior can be represented using a parametrization that 
is linear in the number of observations. This parametrization becomes useful for denser models, 
i.e.~for models that have a larger number of inducing variables.
\item \emph{Extensive experimentation:}
We evaluate \savigp with experiments on small datasets,
medium-scale datasets and two large datasets. 
The experiments on small datasets ($N < 1,300$) 
evaluate  the method under different likelihood models and 
sparsity levels (as determined by the number of inducing variables),  
including problems such as   regression, classification, Log Gaussian Cox processes, 
and warped \gptext{s} \citep{snelson2003warped}. 
We show that \savigp can perform as well as hard-coded inference methods 
under high levels of sparsity.
 The medium-scale experiments consider binary and multi-class 
 classification using the \mnist dataset ($N=60,000$)  and non-stationary 
 regression under the \gprn model \citep{wilson-et-al-icml-12} using 
 the \sarcos dataset ($N \approx 45,000$). Besides showing the 
 competitiveness of our model for problems at this scale, we analyze 
  the effect of learning the inducing inputs, i.e.~the
 location of inducing variables.  
 In our first large-scale experiment, we study the problem of 
 predicting airline delays (using $N=700,000$), and 
 show that our method is on par with the state-of-the-art approach for 
 scalable \gptext regression \citep{hensmangaussian}, 
 which uses full knowledge of the likelihood model. 
 In our second large-scale experiment, we consider the \mnistlarge dataset, 
 which is an augmented version of \mnist  (containing $N= 8,100,000$ observations). 
 We show that by using this augmented dataset we can improve performance 
 significantly. Finally, in an experiment concerning a non-linear seismic inversion problem, 
 we show that our approach can be applied  easily (without any changes to the inference algorithm) 
 to non-standard machine learning tasks and that our inference method can match closely the solution found by  
 more computationally demanding approaches such as \mcmc. 
\end{itemize}
  
Before describing the family of \gptext models that we focus on, we start 
by relating our work to the previous literature concerning the key inference challenges 
mentioned above, i.e.~scalability and dealing with non-linear likelihoods.


\section{Related Work \label{sec:related}}

As pointed out by \citet{rasmussen-williams-book}, there has been a long-standing 
interest in Gaussian processes with early work  dating back at least 
to the 1880s when Danish astronomer and mathematician T.~N.~Thiel,
concerned with determining the distance between Copenhagen and Lund from astronomical observations,
essentially proposed the first mathematical formulation of Brownian motion 
\citep[see e.g.][for details]{lauritzen1981time}. In geostatistics, \gptext regression is 
known as Kriging \citep[see e.g.][]{cressie1993statistics} where, naturally, it has focused 
on 2-dimensional and 3-dimensional problems. 

While the work of 
\citet{ohagan1978curve} has  been quite influential in applying 
\gptext{s} to general regression problems, the introduction of 
\gptext regression to main-stream machine machine learning  by \citet{williams1996gaussian} sparked 
significant interest in the machine learning community. 
Indeed, henceforth, the  community has taken \gptext models well beyond the
 standard regression setting, addressing other problems such as 
non-stationary and heteroscedastic regression \citep{paciorek2004nonstationary,kersting-et-al-icml-2007,wilson-et-al-icml-12}; %
nonlinear dimensionality reduction  \citep{lawrence-jmlr-2005};
classification \citep{williams1998bayesian,nickisch2008approximations};
multi-task learning \citep{bonilla-et-al-nips-08, yu2008gaussian, alvarez-lawrence-nips-08, wilson-et-al-icml-12};
preference learning  \citep{bonilla-et-al-nips-2010}; and ordinal regression \citep{chu2005gaussian}. 
In fact, their book \citep{rasmussen-williams-book} is the de facto reference in any work related
to Gaussian process models in machine learning.  
As we shall see below, despite all these significant advances, 
most previous work in the \gptext community has focused on addressing  
the scalability and the non-linear likelihood challenges in
isolation. 
\subsection{Scalable Models \label{sec:related-scalability}}
The cubic scaling on the number of observations of Gaussian process models, until very recently, 
has hindered the use of these models in a wider variety of applications.
Work on approaching this problem has ranged 
from selecting informative (inducing) datapoints  from the training data 
so as to facilitate  sparse approximations to the posterior \citep{lawrence2002fast} to 
considering these inducing points as continuous parameters 
and optimizing them within a coherent probabilistic framework \citep{snelson2006sparse}. 

Although none of these methods actually scale to very large datasets as their 
time complexity is $\bigO(\m^2 \n)$, where $\n$ is the number of observations 
and $\m$ is the number of inducing points, unifying such approaches from a  
probabilistic perspective \citep{quinonero2005unifying} has been extremely 
valuable to the community, not only to understand what those methods are actually doing 
but also to develop new approaches to sparse \gptext models. 
In particular, the framework of \citet{titsias2009variational} which has been placed within a 
solid theoretical grounding by  \citet{matthews2016sparse},  has become the  underpinning  machinery
 of   modern scalable approaches to \gptext regression and classification.  
 This has been taken 
one step further by  allowing optimization of 
variational parameters within a stochastic optimization framework \citep{hensmangaussian}, 
hence enabling  the applicability of these inference methods to very large datasets.

Besides variational approaches based on  reverse-KL divergence minimization, other methods 
have adopted different inference engines, based \eg on the minimization of the forward KL divergence, 
such as expectation propagation \citep{hernandez2016scalable}. 
It turns out that all  these methods can be seen from a more general perspective as minimization 
of $\alpha$-divergences, see \citet{bui2017unifying} and references therein. 
 
In addition to  inducing-variable approaches to scalability in \gptext-models, 
 other approaches have exploited the relationship between  infinite-feature linear-in-the-parameters 
 models and \gptext{s}.
   In particular, 
   early work investigated truncated linear-in-the-parameters models 
   as approximations to \gptext regression \citep[see e.g.][]{tipping2001sparse}.
  This idea has been developed in an alternative direction that 
  exploits the relationship between a covariance function 
  of a stationary process and its spectral density, hence providing random-feature approximations 
to the covariance function of a \gptext, similarly to the work of 
  \citet{rahimi-recht-nips-2007,rahimi-recht-nips-2008}, who focused on non-probabilistic kernel machines. 
  For example, \citet{lazaro2010sparse}, \citet{gal2015improving} and  \citet{yang-et-al-aistats-2015}
  have followed these types of approaches, with the latter 
  mostly concerned about having fast kernel evaluations. 
  
  Unlike our work, none of these  approaches deals with the harder task 
  of developing scalable inference methods for multi-output problems and 
  general likelihood models. 

\subsection{Multi-output and Multi-task Gaussian Processes}
Developing multi-task or multi-output learning approaches using Gaussian processes has also 
proved an intensive area of research. Most \gptext-based 
models for these problems assume that  correlations between the  outputs are 
induced via linear combinations of  a set of independent latent processes.
Such  linear combinations can use  fixed coefficients \citep[see e.g.][]{teh-et-al-aistats-05}  
or input-dependent coefficients \citep{wilson-et-al-icml-12,nguyen2013efficient}.   
Tasks dependencies can also be defined explicitly through the covariance 
function as done by \citet{bonilla-et-al-nips-08}, who assume a covariance 
decomposition between tasks and inputs.  
More complex dependencies can be modeled  by using the convolution formalism
 as done in earlier work by \citet{boyle-frean-nips-05} 
 and generalized by \citet{alvarez-lawrence-nips-08,alvarez2010efficient},
 whose    later work also provides efficient inference algorithms for such models
\citep{alvarez2011computationally}. Finally, the work of \citet{nguyen-bonilla-uai-2014} also 
assumes a linear combination of independent latent processes but it exploits 
the developments of \citet{hensmangaussian} to scale up to very large datasets. 

Besides the work of \citet{nguyen-bonilla-uai-2014}, these approaches do not scale 
to a very large number of observations and all of them are mainly concerned with regression 
problems. 
 
 \subsection{General Nonlinear Likelihoods}
 The problem of inference in models with \gptext priors and nonlinear likelihoods 
  has been tackled from a sampling perspective 
   with algorithms such as elliptical slice sampling \citep[\ess,][]{murray2009elliptical},
   which  are more effective at drawing samples from strongly correlated Gaussians
   than generic \mcmc algorithms. Nevertheless,
   as we shall see in Section \ref{sec:expts-lgcp}, the sampling cost of \ess 
   remains a major challenge for practical usages.
   
  From a variational inference perspective, the work by \citet{opper-arch-nc-2009}  
 has  been slightly
 under-recognized by the community even though it 
proposes an efficient full Gaussian 
posterior approximation for \gptext models with \iid observations. Our work 
pushes this breakthrough further by allowing multiple latent functions, multiple
outputs, and more importantly, scalability to large datasets. 

Another  approach to deterministic approximate inference  is the integrated 
nested Laplace approximation \citep[\inla,][]{rue2009approximate}.
\inla uses numerical integration to approximate the marginal likelihood, 
which makes it unsuitable for \gptext models that  contain a large number of hyperparameters.

%
%
 \subsection{More Recent Developments \label{sec:related-recent}} 
As mentioned in \S \ref{sec:intro},  the scalability 
 problem  continues to attract the interest of researchers 
 working with \gptext models, with 
 recently developed distributed inference frameworks  \citep{gal-et-al-nips-2014,deisenroth-ng-icml-2015}, 
 and the  variational inference
 frameworks for scalable \gptext regression and classification by 
 \citet{hensmangaussian} and \citet{hensman-et-al-aistats-2015}, respectively. 
 As with the previous work described in \S \ref{sec:related-scalability}, these approaches 
 have been limited to classification or regression problems or specific to a particular class of 
 likelihood models such as \gptext-modulated Cox processes \citep{john2018large}.
 
Contemporarily to the work of \cite{nguyen-bonilla-nips-2014}, which underpins our approach, 
\citet{ranganath2014black} developed black-box variational inference (\bbvi) 
for general latent variable models. 
Due to this generality, it under-utilizes the rich amount of information available 
in Gaussian process models. For example, \bbvi approximates the KL-divergence 
term of the evidence lower bound  but this is  computed analytically in our method. 
Additionally, for practical reasons, \bbvi 
imposes the variational distribution to fully factorize over the latent variables, 
while we make no such a restriction.
A clear disadvantage of \bbvi is that it does not provide a  practical way 
of learning the covariance hyperparameters of \gptext{s} --- in fact, these are set to fixed values.
In principle, these values can be learned in \bbvi using stochastic optimization, 
but experimentally, we have found this to be problematic, ineffectual, and time-consuming.

Very recently, \citet{bonilla2016extended} have used the random-feature approach 
mentioned in Section \ref{sec:related-scalability} along with linearization techniques to 
provide scalable methods for inference in \gptext models with general likelihoods.
Unlike our approach for estimating expectations of the conditional likelihood, 
such linearizations of the conditional likelihood are approximations and generally do not converge to the exact 
expectations even in the limit of a large number of observations. 

Following the recent advances in making automatic differentiation widely available and 
easy to use in practical systems \citep[see e.g.][]{baydin2015automatic}, developments 
on stochastic variational inference for fairly complex probabilistic models 
\citep{rezende2014stochastic,kingma2013auto}  can be used when the likelihood 
can be implemented in such systems and can be expressed using the so-called \emph{re-parametrization 
trick} \citep[see e.g.][for details]{kingma2013auto}. Nevertheless, the advantage of our method over  
these generic approaches is twofold. Firstly, as with \bbvi, in practice such a generality comes at 
the cost of making assumptions 
about the posterior (such as factorization), which is not suitable for \gptext models.
Secondly, and more importantly, such methods are not truly black-box as they require
explicit access to the implementation of the conditional likelihood.  
An interesting example where one cannot apply the re-parametrization trick is given by \cite{challis-barber-jmlr-2013}, who describe a large class of functions (that include the  Laplace log likelihood) that are neither differentiable or continuous but their expectation over a Gaussian posterior is smooth. 
A more general setting where a truly black-box approach is required concerns inversion problems \citep{tarantola-book-2004} where latent functions are passed through domain-specific forward models followed by a known noise model \citep{bonilla2016extended,reid-et-al-ijcai-2013}. These forward models may be non-differentiable, given as an executable, or too complex to re-implement quickly in an automatic differentiation framework. To illustrate this case, we present results on a seismic inversion application in \S \ref{sec:seismic}.  

Nevertheless, we acknowledge that some of these developments mentioned above have been extended to the  \gptext literature, where the {re-parametrization trick} has been used \citep{krauth-et-al-uai-2017,GPflow2017,hensman-et-al-jmlr-2017,pmlr-v70-cutajar17a}. These 
contemporary works show that it is worthwhile building  more tailored methods that may require a lower number 
of samples to estimate the expectations in our framework.

Finally, a related area of research is that of modeling complex data with deep belief networks based on Gaussian process mappings \citep{damianou-lawrence-aistats-2013}, 
which have  been proposed primarily as hierarchical extensions of the Gaussian  process latent variable model  \citep{lawrence-jmlr-2005}. 
Unlike our approach, these models target the unsupervised problem of discovering structure in high-dimensional data and have focused mainly  
on  small-data applications. However, much like the recent work on ``shallow"-\gptext architectures, inference in these models have also been made scalable for supervised and unsupervised learning, exploiting the reparameterization trick, \eg using random-feature expansions  \citep{pmlr-v70-cutajar17a} or inducing-variable approximations \citep{salimbeni2017doubly}.

\section{Latent Gaussian Process Models \label{sec:model}}
Before starting our description of the types of models we are addressing in this paper,
we refer the reader to  \apref{app:notation} for a summary of the notation used henceforth. 
 
We consider supervised learning problems where we are given a 
dataset  $\calD = \{ \x_n, \y_n\}_{n=1}^\n$, where $\x_n$ is a $\d$-dimensional
input vector and $\y_n$ is a $\p$-dimensional output. Our  goal 
is to learn the mapping from inputs to outputs, which can be established 
via $\q$ underlying latent functions $\{f_j\}_{j=1}^{\q}$. 
In many problems these latent functions  have a physical  interpretation, while in others they  
are simply nuisance parameters and we just want to integrate them out in order to make probabilistic predictions. 

A sensible modeling approach to the above problem is to assume that the $\q$ latent functions $\{ f_j \}$ are 
uncorrelated a priori and that they are drawn from $\q$ zero-mean Gaussian processes \citep{rasmussen-williams-book}:
\begin{align}
\pcond{f_j}{\hyperparam_j} & \sim \gp\left(0, \kernel_j(\cdot, \cdot; \hyperparam_j)\right)\text{,} \quad j=1, \ldots \q \text{,} \quad \text{then} \\
\label{eq:prior}
\pcond{\f}{\hyperparam} &= \prod_{j=1}^\q \pcond{\fj}{\hyperparam_j} = \prod_{j=1}^\q \Normal(\fj; \vec{0}, \Kxx) \text{,} 
\end{align}
where $\f$ is the set of all latent function values; $\fj = \{f_j(\x_n)\}_{n=1}^N$ denotes the values of 
latent function $j$;  $\Kxx$ is the covariance matrix 
induced by the covariance function $\kernel_j(\cdot,\cdot; \hyperparam_j)$ 
evaluated at every pair of inputs; and
$\hyperparam = \{ \hyperparam_j \}$ are the parameters of the 
corresponding covariance functions.
Along with the prior in Equation \eqref{eq:prior}, we can also assume that our multi-dimensional 
observations $\{\y_n\}$ are \iid given the corresponding set of latent functions $\{ \f_n\}$:
\begin{align}
\label{eq:likelihood}
\plike = \prod_{n=1}^N \pcond{\yn}{\fn, \likeparam} \text{,}
\end{align}
 where $\y$ is the set of all output observations;  $\y_n$ is the $n\mth$ output observation; 
 $\fn = \{f_j(\x_n)\}_{j=1}^Q$ is the set of latent function values which $\y_n$ depends upon;
 and $\likeparam$ are the conditional likelihood parameters. In the sequel, we will refer
 to the covariance parameters $(\hyperparam)$ as the model hyperparameters. 
 
In other words, we are interested in models for which the following criteria are satisfied:
\begin{enumerate}
\item[(i)]
 \emph{factorization of the  prior  over the latent functions}, as specified by Equation \eqref{eq:prior},  and
\item[(ii)]
 \emph{factorization of the conditional likelihood  over the observations given the latent functions}, 
 as specified by Equation \eqref{eq:likelihood}. 
\end{enumerate}
We refer to the models satisfying the above assumptions (when using \gptext priors) as 
latent Gaussian process models (\lgpm{s}).
Interestingly, a large class of problems can be well modeled  with the above assumptions.   
For example binary classification \citep{nickisch2008approximations,williams1998bayesian}, warped 
\gptext{s} \citep{snelson2003warped}, log Gaussian Cox processes \citep{moller1998log}, 
multi-class classification \citep{williams1998bayesian}, and multi-output regression \citep{wilson-et-al-icml-12} all 
belong to this family of models. 

More importantly, besides the \iid assumption,  
there are not additional constraints on the conditional likelihood  
which can be any linear or nonlinear model. 
Furthermore, as we shall see in \S \ref{sec:inference}, our proposed inference algorithm 
only requires evaluations of this likelihood model in a black-box manner, 
i.e.~without requiring detailed knowledge of its implementation or its gradients.
\subsection{Inference Challenges}
As mentioned in \S \ref{sec:intro}, for general \lgpm{s}, the inference problem of 
estimating the posterior distribution over the latent functions given the 
observed data $\pcond{\f}{\calD}$ and, ultimately, for a new observation $\xstar$,
 estimating the predictive posterior distribution $\pcond{\fstar}{\xstar, \calD}$,  
 poses two important challenges (even in the case of a single latent 
function $\q=1$) from the computational and statistical perspectives. 
These challenges are (i) scalability to large datasets and (ii) dealing with nonlinear likelihoods.

We address the \emph{scalability challenge}  inherent to \gptext models
(given their cubic time complexity on the number of observations) 
 by augmenting our priors using an inducing-variable approach 
 \citep[see e.g.][]{quinonero2005unifying} and embedding our model into a 
variational inference framework \citep{titsias2009variational}. 
In short, inducing variables in sparse \gptext models act as latent summary
statistics, avoiding 
the computation of large inverse covariances.  Crucially, 
unlike other approaches \citep{quinonero2005unifying}, 
we  keep an explicit representation of these variables (and integrate them out 
variationally), 
which facilitates the decomposition of the variational objective 
into a sum on the individual datapoints. This  allows us  
to devise stochastic optimization strategies and 
parallel implementations in cloud computing
services such as  Amazon EC2. 
Such a strategy,  also allows us to learn the location of the
 inducing variables (\ie the inducing inputs) in conjunction with
 variational parameters and  hyperparameters, which in general provides better
performance, especially in high-dimensional problems. 

To address the \emph{nonlinear  likelihood challenge}, which from a 
variational perspective boils down to estimating expectations of a nonlinear function 
over the approximate posterior,  we   
follow a stochastic estimation approach, in which we develop 
low-variance Monte Carlo estimates of the expected log-likelihood term 
in the variational objective. 
Crucially, we will show that  the expected log-likelihood 
term can be estimated efficiently by using only samples from univariate 
Gaussian distributions.

\section{Scalable Inference \label{sec:inference}}
Here we describe our scalable inference method 
 for the model class specified in \S \ref{sec:model}. We build upon
 the inducing-variable formalism underlying most sparse \gptext approximations  
 \citep{quinonero2005unifying,titsias2009variational} and obtain an algorithm 
with time complexity  $\bigO(\m^3)$, where $\m$ is the number of inducing 
 variables per latent process. 
 We show that, under this sparse approximation and 
 the variational inference framework, 
 the expected log likelihood term and its gradient can be estimated using only 
 samples from univariate Gaussian distributions.
 \subsection{Augmented Prior \label{sec:augmented-prior}}
 In order to make inference scalable  we redefine our prior  in terms of some auxiliary variables  
  $\{ \uj \}_{j=1}^\q$, which we will refer to as the \emph{inducing variables}. 
These inducing variables lie in the same space as $\{ \fj\}$ and are drawn from the same zero-mean \gptext  
 priors.  As before, we assume factorization of the prior across the $\q$ latent functions.
 Hence the resulting  augmented prior is given by:
\begin{align}
	\label{eq:prior-u}
	p(\u) & 
	= \prod_{j=1}^\q \Normal(\uj; \vec{0}, \Kzz) \text{,} \qquad
	p(\f | \u) = \prod_{j=1}^\q \Normal(\fj; \priormean, \priorcov) \text{, where }\\
	\priormean &= \Kxz \Kzzinv \uj  \text{, and }\\
	\label{eq:Ktilde-A}
	\priorcov &= \Kxx - \Aj \Kzx \text{ with }
	\Aj  = \Kxz \Kzzinv  \text{,}
\end{align}
where $\uj$ are the inducing variables for latent process $j$;
$\u$ is the set of all the inducing variables;
$\Zj$ are all the inducing inputs for latent process $j$;
$\X$ is the matrix of all input locations $\{\x_n\}$; and
$\Gram{\vec{u}}{\vec{v}}$ is the covariance matrix induced by evaluating 
the covariance function $\kernel_j(\cdot, \cdot)$ at all 
pairwise columns of matrices $\mat{U}$ and $\mat{V}$.
We note that while each of the inducing variables in $\uj$
lies in the same space as the elements in $\fj$, each of the $M$ inducing inputs in $\Zj$
lies in the same space as each input data point $\x_n$.
Therefore, while $\uj$ is a $\m$-dimensional vector, $\Zj$ is a $\m \times \d$ matrix 
where each of the rows corresponds to a $\d$-dimensional inducing input.  
We refer the reader to \apref{app:notation} for a summary of the notation 
and dimensionality of the above kernel matrices. 

As thoroughly investigated by \citet{quinonero2005unifying}, most \gptext approximations
 can be formulated using the augmented prior above and additional assumptions 
 on the training and test conditional distributions $p(\f|\u)$ and $p(\fstar|\u)$, respectively.
 Such approaches have been traditionally referred to as \emph{sparse} approximations and we 
 will use this terminology  as well. Analogously, we will refer to models with a larger number 
 of inducing inputs as \emph{denser} models. 
 
 It is important to emphasize that the joint prior  $p(\f, \u)$ defined in Equation \eqref{eq:prior-u}  
is an equivalent  prior to that in the original model,  as if we integrate 
out the inducing variables $\u$ from this joint prior  we will obtain   
 the  prior $p(\f)$ in Equation \eqref{eq:prior} exactly. 
Nevertheless, as we shall see later, following a variational inference 
approach and having an explicit representation 
of the (approximate) posterior over the inducing variables will 
be fundamental to scaling up inference in these  types of models without making 
the assumptions on the training or test conditionals described in
 \citet{quinonero2005unifying}.

Along with the joint prior defined above, we maintain the 
factorization assumption of the conditional likelihood given 
in Equation \eqref{eq:likelihood}.
\subsection{Variational Inference and the Evidence Lower Bound}
Given the prior in Equation \eqref{eq:prior-u} and the likelihood in Equation \eqref{eq:likelihood}, 
posterior inference for general (non-Gaussian) likelihoods is analytically intractable. Therefore, 
we resort to approximate methods such as variational inference \citep{jordan1998introduction}.
Variational inference methods entail positing a tractable family of distributions and finding the member of the family that is 
``closest" to the true posterior in terms of their  the Kullback-Leibler divergence. 
In our case, we are seeking to approximate the joint posterior  $p(\f,\u | \y)$ with a 
variational distribution $\qjoint$.
\subsection{Approximate Posterior}
Motivated by the fact that the true joint posterior is given by
$	p(\f,\u | \y) = p(\f|\u,\y) p(\u | \y)$, 
our  approximate posterior has the form:
\begin{equation}
\label{eq:joint-posterior}
 \qjoint = p(\f | \u) \qu \text{,}	
\end{equation}
where $p(\f | \u)$ is the conditional prior given in Equation \eqref{eq:prior-u} and 
$\qu$ is our approximate (variational) posterior. This decomposition has proved 
effective in regression problems with a single latent process and a single output 
\citep[see e.g.][]{titsias2009variational}. 

Hence, we can define our variational distribution using a mixture of Gaussians (\mog):
\begin{equation}
\label{eq:var-posterior}
\qu = \sum_{k=1}^{\k} \pi_k q_k(\u | \postmean{k}, \postcov{k}) = \sum_{k=1}^{\k} \pi_k \prod_{j=1}^\q  \Normal(\uj; \postmean{kj}, \postcov{kj}) \text{,}
\end{equation}
where $\llambda = \{ \pi_k, \postmean{kj}, \postcov{kj} \}$ are the variational parameters: 
the mixture proportions $\{ \pi_k \}$,  
the posterior means  \{$\postmean{kj}\}$ and posterior covariances $\{\postcov{kj}\}$ of 
the inducing variables corresponding 
to mixture component $k$ and latent function $j$. 
We also note that each of the mixture components 
$q_k(\u | \postmean{k}, \postcov{k})$ is a Gaussian with mean $\postmean{k}$ 
and block-diagonal covariance $\postcov{k}$. 

An early reference for using a mixture of Gaussians (\mog) within variational inference is given by \citet{bishop-et-al-nips-1998} in the context of Bayesian networks. Similarly, \citet{gershman-et-al-icml-12} have used \mog for non-\gptext models and, unlike our approach, used a second-order Taylor series approximation of the variational lower bound. 
\subsection{Variational Lower Bound}
It is easy to show that minimizing the Kullback-Leibler divergence between
our approximate posterior and the true posterior, $\kl{\qjoint}{p(\f, \u | \y)}$ is 
equivalent to maximizing the log-evidence lower bound ($\elbo$), which is composed 
of a KL-term ($\klterm$) and an expected log-likelihood term ($\ellterm$). In other words:
\begin{align}
\label{eq:elbo}
\log p(\y) 
\geq   
\elbo(\llambda)  
&
\defeq
\klterm(\llambda) + \ellterm(\llambda) \text{, where } \\
\label{eq:kl}
\klterm(\llambda) 
& =
- \kl{\qjoint}{p(\f, \u)} \text{, and}\\
\label{eq:ell}
\ellterm(\llambda)
&=
 \expectation{\qjoint}{\log p(\y | \f)}
\text{,}
\end{align}
where $\expectation{q(x)}{g(x)}$ denotes the expectation of function $g(x)$ over distribution $q(x)$. 
Here we note that $\klterm$ is a negative KL divergence between the joint approximate 
posterior $\qjoint$ and the joint prior $p(\f, \u)$. 
Therefore, maximization of the $\elbo$ in Equation \eqref{eq:elbo} entails minimization of 
an expected loss (given by the negative 
expected log-likelihood $\ellterm$) regularized by  $\klterm$, which imposes 
the constraint of finding solutions to our approximate posterior 
that are close to the prior in the KL-sense.
 
An interesting observation of the decomposition of the $\elbo$ objective is that,
unlike $\ellterm$, 
$\klterm$  in Equation \eqref{eq:kl} does not depend on the 
conditional likelihood $p(\y | \f)$, for which we do not assume 
any specific parametric form (i.e.~black-box likelihood).
We can thus address the technical difficulties regarding each component and their 
gradients  separately using different approaches.
In particular, for $\klterm$ we will exploit the structure of the variational posterior 
in order to avoid computing KL-divergences over distributions involving all the 
data. Furthermore, we will  obtain a lower bound for $\klterm$ in the  
general case of  $q(\u)$ being a mixture-of-Gaussians (\mog) as given 
in Equation \eqref{eq:var-posterior}. For the 
expected log-likelihood term ($\ellterm$) we will use 
a Monte Carlo approach in order to estimate the required expectation and 
its gradients.
\subsection{Computation of the KL-divergence term ($\klterm$)}
In order to have an explicit form for $\klterm$ and its gradients, we start by 
expanding Equation  \eqref{eq:kl}:
\begin{align}
\label{eq:kl1}
\klterm(\llambda) 
& =- \kl{\qjoint}{p(\f, \u)}  
=  - \expectation{\qjoint} {\log \frac{\qjoint}{p(\f, \u)}} \text{,} \\
\label{eq:kl2}
&=  - \expectation{p(\f | \u) \qu}{\log \frac{q(\u |  \llambda)}{p(\u)}} \text{,} \\
\label{eq:kl3}
&= - \kl{\qu}{p(\u)} \text{,}
\end{align}
where we have applied the definition of the KL-divergence in Equation 
\eqref{eq:kl1};  used the variational joint posterior $\qjoint$ given in Equation 
\eqref{eq:joint-posterior} to go from Equation  \eqref{eq:kl1} to Equation \eqref{eq:kl2}; 
and  integrated out $\f$ to obtain Equation \eqref{eq:kl3}.
We note that the definition of the joint posterior $\qjoint$  in Equation 
\eqref{eq:joint-posterior} has been crucial to transform a KL-divergence between
the joint approximate posterior and the joint prior into a KL-divergence 
between the variational posterior $\qu$ and the prior $p(\u)$ over the inducing 
variables. In doing that, we have avoided computing  a KL-divergence between 
distributions over the $\n$-dimensional variables $\fj$. As we shall see later, 
this implies a reduction in  time complexity  from $\bigO(\n^3)$ to
$\bigO(\m^3)$, where $\n$ is the number of datapoints and $\m$ is the 
number of inducing variables. 

We now decompose the resulting KL-divergence term in Equation \eqref{eq:kl3} as follows,
\begin{align}
\klterm(\llambda) &= - \kl{\qu}{p(\u)} = \enterm(\llambda) + \crossterm(\llambda) \text{, where:} \\
\enterm(\llambda) &=  - \expectation{\qu}{\log \qu} \text{, and} \\
\label{eq:lcross}
 \crossterm(\llambda) &= \expectation{\qu}{\log p(\u)} \text{,}
\end{align}
where $\enterm(\llambda)$ denotes the differential entropy of the approximating distribution $\qu$ and 
$\crossterm(\llambda)$ denotes the negative cross-entropy between the approximating distribution $\qu$ and
the prior $p(\u)$. 

Computing the entropy of the variational distribution in Equation \eqref{eq:var-posterior}, which is a mixture-of-Gaussians (\mog),
is analytically intractable. 
However, a lower bound of this entropy can be obtained using Jensen's inequality \citep[see e.g.][]{ihuber-et-al-2008} giving:
\begin{equation}
\label{eq:entropyhat}
 \enterm(\llambda) \geq 
 - \sum_{k=1}^{\k} \pi_k \log \sum_{\ell=1}^\k \pi_{\ell} \Normal(\postmean{k}; \postmean{\ell}, \postcov{k} + \postcov{\ell}) 
	\defeq \entermhat \text{.}
\end{equation}
The negative cross-entropy in Equation \eqref{eq:lcross} between a Gaussian mixture $\qu$ and 
a Gaussian $p(\u)$, can be obtained analytically, 
\begin{equation}
\label{eq:crossentropy}
\crossterm(\llambda) = - \frac{1}{2} \sum_{k=1}^\k \pi_k \sum_{j=1}^\q [\m \log 2 \pi + \log \det{\Kzz} 
	+ \postmean{kj}^T \Kzzinv \postmean{kj} + \trace{\Kzzinv \postcov{kj}}] \text{.}	
\end{equation}
The derivation of the entropy term and the cross-entropy term in 
Equations  \eqref{eq:entropyhat} and \eqref{eq:crossentropy} is given in \apref{app:klterm}. 
\subsection{Estimation of the expected log likelihood term ($\ellterm$) \label{sec:ell}}
We now address the computation of the expected log likelihood term in Equation \eqref{eq:ell}. 
The main difficulty of computing this term is that, unlike the $\klterm$ where we have full knowledge of 
the prior and the approximate posterior, here we do not assume a specific form for the  conditional likelihood 
$\plike$. Furthermore, we only require evaluations of  $\log\pcond{\yn}{\fn, \likeparam}$ 
for each datapoint $n$, hence  yielding a truly black-box likelihood method. 

We will show one of our main results, 
that of \emph{statistical efficiency} of our $\ellterm$ estimator. This means that, despite having a full 
Gaussian approximate posterior, estimation of the   $\ellterm$ and its gradients only requires 
samples from univariate Gaussian distributions. 
We start by expanding Equation $\eqref{eq:ell}$ using our 
definitions of the approximate posterior and the factorization of the conditional likelihood:
\begin{equation}
\label{eq:ellterm-original}
\ellterm(\llambda)
=
 \expectation{\qjoint}{\log \plike}   =  \expectation{\qf}{\log \plike} \text{,} \\ 
\end{equation}
where, given the definition of the approximate joint posterior $\qjoint$ in Equation \eqref{eq:joint-posterior},
the distribution $\qf$ resulting from marginalizing $\u$ from this joint posterior can be obtained analytically,
\begin{align} 
 	\label{eq:qf}
	\qf &= \sum_{k=1}^\k \pi_k \qkf =  \sum_{k=1}^\k \pi_k \prod_{j=1}^\q \Normal(\fj; \qfmean{kj}, \qfcov{kj}) \text{, with}\\
	\label{eq:meanqf}
	\qfmean{kj} & = \Aj \postmean{kj} \text{, and } \\
	\label{eq:covqf}
	\qfcov{kj} &= \priorcov + \Aj \postcov{kj} \Aj^T \text{,}
\end{align}
where  $\priorcov$ and $\Aj$ are given in Equation \eqref{eq:Ktilde-A} and, 
as defined before,  \{$\postmean{kj}\}$ and  $\{\postcov{kj}\}$ are the posterior means and posterior covariances
of the inducing variables corresponding  to mixture component $k$ and latent function $j$. 
We are now ready to state our result of a statistically efficient estimator for the $\ellterm$:
\newtheorem{theorem1}{Theorem}
\begin{theorem1}
\label{th:univariate}
For the  \gptext model with prior defined in Equations \eqref{eq:prior-u}
to \eqref{eq:Ktilde-A}, and conditional likelihood defined in Equation \eqref{eq:likelihood},
the expected log likelihood over the variational distribution in 
Equation \eqref{eq:joint-posterior} and its gradients can
be estimated using samples from univariate Gaussian distributions.
\end{theorem1}
The proof is constructive and can be found in  \apref{app:proof-th-univariate}. 
We note that a less general result, for the case of one latent function and a single variational Gaussian posterior, 
was obtained in \cite{opper-arch-nc-2009} using a different derivation. 
Here we 
state our final result on how to compute these estimates:
\begin{align}
\label{eq:final-ell}
\ellterm(\llambda) 
&= \sum_{n=1}^\n \sum_{k=1}^\k \pi_k \expectation{\qknf}{ \log \pcond{\yn}{\fn, \likeparam} }  \text{,} \\
\label{eq:ellgradients-1}
\grad_{\llambda_k}  \ellterm(\llambda) 
&=  \pi_k  \sum_{n=1}^\n \expectation{\qknf} {\grad_{\llambda_k}  \log \qknf  \log \pcond{\yn}{\fn, \likeparam}  }
\text{, for } \llambda_k \in \{\postmean{k}, \postcov{k} \} \text{,}  \\
\label{eq:ellgradients-2}
\grad_{\pi_k}  \ellterm(\llambda) 
&=  \sum_{n=1}^\n \expectation{\qknf} { \log \pcond{\yn}{\fn, \likeparam}  } \text{,}
\end{align}
where $\qknf$ is a $\q$-dimensional Gaussian with:
\begin{equation}
		\qknf = \Normal(\fn; \qfmean{k(n)}, \qfcov{k(n)}) \text{,}
\end{equation}
where  $\qfcov{k(n)}$ is a \emph{diagonal} matrix. The $j\mth$ element of the mean and the $(j,j)\mth$ entry of the 
covariance of the above distribution  are given by:
\begin{equation}
\label{mean-cov-postqf}
\vecentry{ \qfmean{k(n)} }{ j } = \ajn^T \postmean{kj}  \text{,} \qquad
\matentry{ \qfcov{k(n)} }{j}{j} =  \matentry{ \priorcov }{n}{n} + \ajn^T \postcov{kj} \ajn \text{,}
\end{equation}
where  $\ajn \defeq  \matcol{\Aj}{n}$   denotes the $\m$-dimensional vector corresponding to the $n\mth$ column of matrix
 $\Aj$; $\priorcov$  and $\Aj$ are given in Equation \eqref{eq:Ktilde-A}; and,
 as before, $\{\postmean{kj},  \postcov{kj} \}$ are the variational parameters corresponding to 
 the mean and covariance of the approximate posterior over the inducing variables for 
 mixture component $k$ and latent process $j$.
 
We emphasize that when $\q > 1$, $\qknf$ is not a univariate marginal but a $\q$-dimensional 
marginal posterior with diagonal covariance. Therefore, only samples from univariate Gaussians are required to 
estimate the expressions in Equations \eqref{eq:final-ell} to  \eqref{eq:ellgradients-2}.  
\subsubsection{Practical consequences and unbiased estimates} 
There are two immediate practical consequences of the result in Theorem \ref{th:univariate}.
The first consequence is that we can use unbiased empirical estimates of the expected log likelihood term 
and its gradients. In our experiments we use  Monte Carlo (\mc) estimates, hence we can compute $\ellterm$  as:
\begin{align}
	\left\{ \fn^{(k,i)} \right\}_{i=1}^{S} &\sim \Normal(\fn; \qfmean{k(n)}, \qfcov{k(n)} ) \text{, }  k=1,\ldots, \k  \text{,}\\
	\label{eq:elltermhat}
	\elltermhat &=  \frac{1}{\s} \sum_{n=1}^\n \sum_{k=1}^\k \pi_k \sum_{i=1}^{\s} \log p(\yn | \fn^{(k,i)}, \likeparam) \text{,}
\end{align}
where $\qfmean{k(n)}$ and $\qfcov{k(n)}$  are the vector and matrix forms of the mean and covariance 
of the posterior over the latent functions as given in Equation \eqref{mean-cov-postqf}.  
Analogous \mc  estimates of the gradients are given in Appendix  \ref{app:mc-grad-ell}.  

The second practical consequence is that in order to estimate the gradients of the 
$\ellterm$, using Equations \eqref{eq:ellgradients-1} and \eqref{eq:ellgradients-2},
we only require evaluations of the conditional likelihood in a black-box manner, without resorting
to numerical approximations or analytical derivations of its gradients.

\section{Parameter Optimization \label{sec:optimization}}
In order to learn the parameters of our model we seek to maximize our (estimated) 
log-evidence lower bound ($\elbohat$) using gradient-based optimization. 
Let $\param = \{\llambda, \hyperparam,\likeparam\}$ be all the model 
 parameters  
for which point estimates are required. We have that: 
\begin{align}
\elbohat (\param)
& \defeq \entermhat (\param)  + \crossterm(\param)   + \elltermhat (\param) \text{,} \\
\grad_{\param}\elbohat
&=
\grad_{\param}\entermhat + \grad_{\param} \crossterm + \grad_{\param}  \elltermhat  \text{,} 
\end{align} 
where we have made explicit the  
dependency of the log-evidence lower bound on 
any parameter of the model and 
$\entermhat$, $\crossterm$, and $\elltermhat$ are given in Equations
\eqref{eq:entropyhat}, \eqref{eq:crossentropy}, \eqref{eq:elltermhat} respectively.
The gradients of $\elbohat$ wrt variational parameters $\llambda$, 
covariance hyperparameters $\hyperparam$ and likelihood parameters
$\likeparam$ are given in Appendices \ref{app:gradients-variational}, 
\ref{app:grads-hyper}, and \ref{app:grads-like}, respectively. 
As shown in these appendices, not all constituents of $\elbohat$ contribute to 
learning all parameters, for example 
$\grad_{\hyperparam}\entermhat = 
\grad_{\likeparam}\entermhat =
 \grad_{\likeparam}\crossterm=0$. 

Using the above objective function ($\elbohat$) and its gradients we can consider 
batch optimization with limited-memory \bfgs for small datasets 
and medium-size  datasets. 
However, under this batch setting,  the computational 
cost can be too large to be afforded in practice even for medium-size datasets 
on single-core architectures. 
\subsection{Stochastic optimization} 
To deal with the above problem, we first note that the
terms corresponding to the KL-divergence $\entermhat$ and $\crossterm$ in Equations 
\eqref{eq:entropyhat} and \eqref{eq:crossentropy} do not depend on the observed data, 
hence their computational complexity is independent of $\n$. 
More importantly, we  note that  $\elltermhat$ in 
Equation \eqref{eq:elltermhat} decomposes as a sum of  expectations over 
 individual datapoints.  This makes our inference framework amenable to 
parallel computation and stochastic optimization \citep{robbins1951stochastic}.
More precisely, we can rewrite  $\elbohat$ as:
\begin{equation}
	\elbohat = \sum_{n=1}^{\n} \left[ \frac{1}{\n} \left( \entermhat + \crossterm \right)
	+ \elltermhatn \right] \text{,}
\end{equation}
where $\elltermhatn = \frac{1}{\s}  \sum_{k=1}^\k \pi_k \sum_{i=1}^{\s} \log p(\yn | \fn^{(k,i)}, \likeparam)$, which enables us to apply stochastic optimization 
techniques such as stochastic gradients descend \citep[\sgd,][]{robbins1951stochastic}
or \adadelta \citep{zeiler2012adadelta}. The complexity of the 
resulting algorithm is independent of $\n$ and 
dominated by algebraic operations that are 
 $\bigO(\m^3)$ in time,  where $\m$ is the number of inducing points per latent process.  
This makes our  automated variational inference framework practical for very large 
 datasets. 
 \subsection{Reducing the variance of the gradients with control variates}
 Theorem \ref{th:univariate} is fundamental to having a statistical efficient algorithm that 
 only requires sampling from univariate Gaussian distributions (instead of sampling 
 from very high-dimensional Gaussians) for the estimation of the expected log likelihood 
 term and its gradients. 
 
 However, the variance of the gradient estimates may be too large 
 for the algorithm to work in practice, and variance reduction techniques become necessary. 
Here we use the well-known technique of control variates \citep[see e.g.][\secrefout{8.2}]{ross2006simulation},
where a new gradient estimate is constructed so as to have the same expectation 
but lower variance than the original estimate.  Our control variate is the 
so-called score function $h(\fn) =  \grad_{\lambda_k} \log \qknf$  
and full details are given in \apref{app:ctrl-variates}. 
 \subsection{Optimization of the inducing inputs}
 So far we have discussed optimization of variational parameters ($\varparam$), 
 i.e.~the parameters of the approximate posterior ($\{\pi_k, \postmean{k}, \postcov{k}\}$); 
 covariance hyperparameters ($\hyperparam$); and likelihood parameters
 ($\likeparam$). However, as discussed by \citet{titsias2009variational}, 
 the inducing inputs $\{\Z_j\}$ can be seen as additional variational parameters and,
 therefore, their optimization should be somehow robust to overfitting. 
 As described in the Experiments (Section \ref{sec:expts-learn-z}),  learning of the inducing inputs 
 can improve performance, requiring a lower number of inducing variables than 
 when these are fixed. This, of course, comes at an additional computational cost 
 which can be significant when considering high-dimensional input spaces. 
 As with the variational parameters, we study learning of the inducing inputs via
 gradient-based optimization, for which we use the gradients provided in 
 \apref{app:grad-Z}. 

Early references in the machine learning community where a single variational objective is used for parameter inference (in addition to posterior estimation over latent variables) can be found in \citet{hinton-vancamp-colt-1993,mackay-snn-1995,lappalainen-ica-2000}. 
These methods are now known under the umbrella term of variational Bayes\footnote{We note that these methods were then referred to as ``Ensemble Learning", since a full distribution (i.e.~an ensemble of parameters) was used instead of a single point estimate. Fortunately, nowadays, variational Bayes is a preferred term since Ensemble Learning is more commonly associated with methods for combining multiple models.} and consider a prior and an approximate posterior for these parameters within the variational framework. As mentioned above, rather than a full variational Bayes approach, we provide point estimates of 
$\{\hyperparam,  \likeparam\}$ and $\{\Zj\}$, and our experiments in \secref{sec:expts} confirm the efficacy of our approach. More specifically, we show in \secref{sec:expts-learn-z} that point-estimation of the inducing inputs $\{\Zj\}$ using the variational objective can be significantly better than using  heuristics such as k-means clustering.

 \section{Dense Posterior and Practical Distributions}
 In this section  we consider the case when the inducing inputs are 
 placed at the training points, i.e.~$\Z_j = \X$ and 
 consequently  $\m=\n$. 
 As mentioned in Section \ref{sec:augmented-prior},
  we refer to this setting as dense  to distinguish it from the case 
  when $\m < \n$, for which the resulting models are usually called sparse approximations.
   It is important to realize that not all 
   real datasets are very large and that in many cases the resulting time and 
   memory complexity $\bigO(\n^3)$ and $\bigO(\n^2)$ can be afforded. 
   Besides the dense posterior case, we also study some 
 particular variational distributions that make our framework more 
 practical, especially in large-scale applications. 
 \subsection{Dense Approximate Posterior}
When our posterior is dense the only approximation made  is the assumed variational distribution 
in Equation \eqref{eq:var-posterior}. We will show 
 that, in this case, we recover the objective function in \citet{nguyen-bonilla-nips-2014} and 
 that hyper-parameter learning is easier as the terms in the 
 resulting objective function that depend on the hyperparameters do not involve 
 \mc estimates. Therefore, their analytical gradients can be used. Furthermore, in the following 
 section, we will provide an efficient exact 
 parametrization of the posterior covariance that reduces the $\bigO(\n^2)$ memory complexity
 to a linear complexity $\bigO(\n)$.
 
 We start by looking at the components of $\elbo$ when we make $\Z_j = \X$, which we can simply 
 obtain by making $\Kzz = \Kxx$ and $\m = \n$,  and realizing that the posterior parameters $\{\postmean{kj}, \postcov{kj} \}$ 
 are now $\n$-dimensional objects. Therefore, we leave the entropy term $\entermhat$ in Equation 
 \eqref{eq:entropyhat} unchanged and we replace all the appearances of $\Kzz$ with $\Kxx$
 and all the appearances of $\m$ with $\n$ for the $\crossterm$. We refer the reader to 
 \apref{app:nonsparse} for details of the equations but it is easy to see that 
 the resulting $\entermhat$ and $\crossterm$ are identical to those 
 obtained by \citet[Equations 5 and 6]{nguyen-bonilla-nips-2014}.  
 
 For the expected log likelihood term, the generic expression in Equation 
 \eqref{eq:ellterm-original} still applies but we need to figure out the resulting 
 expressions for the approximate posterior parameters in Equations
\eqref{eq:meanqf} and \eqref{eq:covqf}. It is easy to show that 
 the resulting posterior means
and covariances are in fact $\qfmean{kj} = \postmean{kj}$
and $\qfcov{kj} = \postcov{kj}$ (see \apref{app:nonsparse} for details). 
This means that in the dense case we simply estimate the $\elltermhat$ by using empirical expectations 
over the unconstrained variational posterior $\qf$, with `free' mean and covariance parameters. 
In contrast, in the sparse case, although these expectations 
are still computed over  $\qf$, the parameters of the variational posterior 
$\qf$ are constrained by Equations \eqref{eq:meanqf} and \eqref{eq:covqf} which 
are functions of the prior covariance and the parameters of the variational distribution over the 
inducing variables $\qu$. As we shall see in the following section, this distinction between 
the dense case and sparse case has critical consequences on hyperparameter learning. 
\subsubsection{Exact Hyperparameter Optimization}
The above insight reveals a remarkable property of the model in the dense case. 
Unlike the sparse case, the expected log likelihood term does not depend 
on the covariance hyperparameters, as the expectation of the conditional likelihood 
is taken over the variational distribution $\qf$ with `free' parameters. 
Therefore,  only  the cross-entropy term $\crossterm$ depends on the hyperparameters
 (as we also know that  $\grad_{\hyperparam}\entermhat=0$). 
 For this term, as seen in \apref{app:nonsparse-elbo} and corresponding 
 gradients in Equation \eqref{eq:grad-cross-hyper},  we have derived the 
 exact (analytical) expressions for the objective  function and its gradients, 
 avoiding empirical \mc estimates altogether. 
 This has a  significant practical implication: despite using black-box inference,  
 the hyperparameters are optimized wrt the true evidence lower bound (given fixed variational parameters). 
This is an additional and crucial advantage of our automated inference method 
over other generic inference techniques  \citep[see e.g.][]{ranganath2014black},
which do not exploit knowledge of the prior. 
 \subsubsection{Exact Solution with Gaussian Likelihoods}
Another interesting property of our approach arises from the fact that, 
as we are using \mc estimates,  $\elltermhat$ is an unbiased 
estimator of $\ellterm$. This means that, as the number of samples $\s$
increases, $\elbohat$ will converge to the true value  $\elbo$. In the 
case of a Gaussian likelihood, the posterior over the latent functions 
is also Gaussian and a variational approach with a full Gaussian 
posterior will converge to the true parameters of the posterior,
 see e.g.~\citet{opper-arch-nc-2009} and \apref{app:gaussian-like} for details.
\subsection{Practical Variational Distributions}  
As we have seen in Section \ref{sec:optimization},  learning of all parameters 
in the model can be done in a scalable way through stochastic optimization for 
general likelihood models, providing automated variational inference for 
models with Gaussian process priors.
However, the general \mog approximate posterior in Equation \eqref{eq:var-posterior}
requires  $\calO(\m^2)$ variational parameters for each covariance matrix of the 
corresponding latent process, yielding a total requirement  of $\bigO(\q \k \m^2)$ parameters.
This may cause difficulties for learning when these  parameters are optimized simultaneously.
In this section we introduce two special members of the assumed variational posterior family 
that  improve the practical tractability of our inference framework. These members are 
a full Gaussian posterior and a mixture of diagonal Gaussians posterior.
\subsubsection{Full Gaussian Posterior}
This instance considers the case of only one component in the mixture ($\k=1$) in 
Equation \eqref{eq:var-posterior},which 
has a Gaussian distribution with a \emph{full covariance} matrix for each latent process.
Therefore, following the factorization assumption in the posterior across latent processes,
the posterior distribution over the inducing variables $\u$, and consequently over the latent 
functions $\f$, is a Gaussian with block diagonal covariance, where each block is a 
full covariance corresponding to that of a single latent function. We thus refer to this 
approximate posterior  as the full Gaussian posterior (\full). 
\subsubsection{Mixture of Diagonal Gaussians Posterior}
 Our second practical variational posterior considers a  mixture distribution as  
 in Equation  \eqref{eq:var-posterior}, constraining each of the mixture components 
 for each latent process to be a Gaussian distribution with \emph{diagonal covariance}
 matrix.  Therefore, 
 following the factorization assumption in the posterior across latent processes,
the posterior distribution over the inducing variables $\u$ is a mixture of diagonal 
Gaussians. However, we note that, as seen in Equations \eqref{eq:meanqf} and 
\eqref{eq:covqf}, the posterior over the latent functions $\f$ is not a mixture of 
diagonal Gaussians in the general sparse case. Obviously, in the dense 
case (where $\Zj = \X$) the posterior covariance over $\f$ of each component  does 
have a diagonal structure. Henceforth, we will refer to this approximation simply as
\modg, to distinguish it from the \full case above, while avoiding the use of additional notation.
One immediate benefit of using this approximate posterior is computational, as 
we avoid the inversion of a full covariance 
for each component in the mixture.

As we shall see in the following sections, there are additional benefits from the assumed 
practical distributions and they concern the efficient parametrization of the covariance 
for both  distributions and the lower variance of the gradient estimates 
for the \mog posterior.
\subsection{Efficient Re-parametrization \label{sec:efficient-param}}
As mentioned above, one of the main motivations for having specific practical distributions  
is to reduce the computational overhead due to the large number of parameters to 
optimize. For the \modg approximation, it is obvious that only  $\bigO(\m)$ 
parameters for each latent process and mixture component are required to represent 
the posterior covariance, hence one obtains an efficient parametrization by 
definition. However, for the full Gaussian (\full) approximation, naively, one would require 
$\bigO(\m^2)$ parameters. The following theorem states that for settings that 
require a large number of inducing variables, the \full variational distribution can
be represented using a significantly lower number of parameters.   
\newtheorem{theorem2}[theorem1]{Theorem}
\begin{theorem2}
\label{th:efficient-param}
The optimal full Gaussian variational posterior can be represented using 
a parametrization that is linear in the number of observations ($N$). 
\end{theorem2}
Before proceeding with the analysis of this theorem, we remind the reader that 
the general form of our variational distribution in Equation \eqref{eq:var-posterior}
requires $\bigO(\k \q \m^2)$ parameters for the covariance, for a model with $\k$ mixture components, 
$\q$ latent processes and $\m$ inducing variables. Nevertheless, for simplicity 
and because usually $\k \ll \m$ and $\q \ll \m$, we will omit $\k$ and $\q$ in  
in the following discussion. 

The proof the theorem can be found in \apref{app:efficient-param}, where it 
is shown that in the \full case
the optimal solution for the posterior covariance is given by:
\begin{equation}
\label{eq:Sopt-main}
\widehat{\postcov{j}} = \Kzz \left(\Kzz + \Kzx  \Lambdaj  \Kxz \right)^{-1} \Kzz \text{,}
\end{equation}
where  $\Lambdaj$ is a $\n$-dimensional diagonal matrix. Since the optimal covariance 
can be expressed in terms of fixed kernel computations and a free set 
of parameters given by $\Lambdaj$, only $\n$ parameters are necessary to 
represent the posterior covariance. As we shall see below, this parametrization 
becomes useful for denser models, i.e.~for models that have a large number of 
inducing variables. 
\subsubsection{Sparse Posterior}
In the sparse case  the inducing inputs $\Z_j$ are at arbitrary locations and, more importantly, $M \ll N$, the number 
of inducing variables is considerably smaller than the number of training points. The result in Equation $\eqref{eq:Sopt-main}$ implies that 
if we parameterize $\postcov{j}$ in that way, we will need $\bigO(\n)$ parameters instead of $\bigO(\m^2)$. 
Of course this is useful when roughly $\n < \frac{M^2}{2}$. 
A natural question arises when we define the number 
of inducing points as a fraction of the number of training points, i.e.~$\m = \sfactor \n$ with $0 \leq \sfactor \leq 1$, when is such a parameterization  useful? 
In this case, using the alternative parametrization will become beneficial when $\sfactor > \sqrt{\frac{2}{N}}$. To illustrate this,
consider for example the \mnist dataset used in our medium-scale experiments in 
Section \ref{sec:expts-medium} where $\n = 60,000$. This yields a beneficial regime 
around roughly $\sfactor > 0.006$, which is a realistic setting.
In fact, in our experiments on this dataset we did consider sparsity factors of this magnitude. For example, our 
biggest experiment used $\sfactor = 0.04$. With a naive parametrization of the posterior covariance, we  
would need roughly $2 \times 10^6$ parameters. In contrast, by using the efficient parametrization we only need $50 \times 10^3$ parameters. As shown below, these gains are greater as the model becomes denser, yielding 
a dramatic reduction in the number of parameters when having a fully dense Gaussian posterior.
\subsubsection{Dense Posterior}  
In the dense case we have that $\Z_j = \X,  \forall j=1, \ldots, Q$ and 
consequently $\m = N$. Therefore we have that the optimal posterior covariance is given by:
\begin{align}
\widehat{\postcov{j}} & =	\left( (\Kxx)^{-1}+  \Lambdaj   \right)^{-1}  \text{.}
\end{align}
In principle, the parametrization of the posterior covariance would require $\bigO(\n^2)$ parameters for each latent process.
However, the above result shows that we can parametrize these covariances efficiently using only $\bigO(\n)$ parameters. 

We note that, for  models with $Q=1$,  this efficient re-parametrization has been used by \citet{sheth2015sparse} in the sparse case and \citet{opper-arch-nc-2009} in the dense case, while adopting  an  inference algorithm different to ours.
\subsection{Automatic Variance Reduction with a Mixture-of-Diagonals Posterior}
An additional benefit of having a mixture-of-diagonals (\modg) posterior in the 
dense case is that  optimization  
of the variational parameters will typically converge faster when using a mixture of diagonal Gaussians. 
This is an immediate consequence of the following theorem.
\newtheorem{theorem3}[theorem1]{Theorem}
\begin{theorem3}
\label{th:modg-lower-variance}
When having a dense posterior, the estimator of the gradients wrt the variational parameters using the mixture of 
diagonal Gaussians has a lower variance than the full Gaussian posterior's.
\end{theorem3}
The proof is in \apref{app:modg-lower-variance} 
 and   is simply a manifestation 
of  the Rao-Blackwellization technique \citep{casella1996rao}. The 
theorem is only made possible due to the analytical tractability of the KL-divergence term ($\klterm$) in the 
variational objective ($\elbo$).
The practical consequence of this theorem is that  optimization will typically converge 
faster when using a mixture-of-diagonals Gaussians than when using a full Gaussian posterior 
approximation.

\section{Predictions}
Given the general posterior over the inducing variables $\qu$
in equation \eqref{eq:var-posterior}, 
 the predictive distribution for a new test point $\xstar$ is given by:
\begin{align}
\nonumber
p(\ystar | \xstar) 
&= \sum_{k=1}^\k \pi_k \int p(\ystar | \fstar) \int p(\fstar | \u) \qku \der \u \der \fstar. \\
\label{eq:pred-distri}
&=  \sum_{k=1}^\k \pi_k \int p(\ystar | \fstar) \qkfstar \der \fstar \text{,}
\end{align}
where $\qkfstar$ is the predictive distribution  over the  $\q$ latent functions 
corresponding to mixture component $k$ given 
the learned variational parameters $\llambda = \{\postmean{k}, \postcov{k} \}$:
 \begin{align}
 	\qkfstar &= \prod_{j=1}^\q \Normal(\fstarj; \predmean{kj}, \predvar{kj} ) \text{, with} \\
	\predmean{kj} & = \kernel_j(\xstar, \Zj) \Kzzinv \postmean{kj} \text{, and } \\
	\predvar{kj}  
	&=  \kernel_j(\xstar, \xstar)  - 
	\kernel_j(\xstar, \Zj) 
	\left( 
	\Kzzinv - \Kzzinv \postcov{kj} \Kzzinv
	\right)\kernel_j( \Zj, \xstar) \text{.}
 \end{align}
Thus, the probability of the test points taking values $\ystar$ (e.g.~in classification) 
in Equation \eqref{eq:pred-distri}
can  be readily estimated via Monte Carlo sampling.

\section{Complexity analysis}
\label{sec:complexity}
\newcommand{\bsize}{B}
\newcommand{\tcost}{T}
\newcommand{\likecost}{L}
Throughout this paper we have only considered computational complexity with respect to the number of
inducing points and training points for simplicity.
Although this has also been common practice in previous work
\citep[see e.g.][]{dezfouli-bonilla-nips-2015,nguyen-bonilla-nips-2014,hensmangaussian,titsias2009variational},
we believe it is necessary to provide an in-depth complexity analysis of the
computational cost of our model. 
Here we analyze the computational cost of evaluating the $\elbo$ and its gradients once. 

We begin by reminding the reader of the dimensionality notation used so far 
and by introducing additional notation specific to this section. 
We analyze the more general case of stochastic optimization using mini-batches.  
Let $\k$ be the number of mixture components in our 
variational posterior;
$\q$ the number of latent functions; 
$\d$ the dimensionality of the input data;
$\s$ the number of samples used to estimate the required 
expectations via Monte Carlo; 
$\bsize$ the number of inputs used per mini-batch;  
$\m$ the number of inducing inputs; and  
$\n$ the number of observations.
 We note that in the case
of batch optimization $\bsize=\n$, hence our analysis applies to both the stochastic and batch setting.
We also let $\tcost(e)$ represent the computational costs of expression $e$. 
Furthermore, we assume that the kernel function is simple enough such that 
evaluating its output between two points is $\bigO(\d)$ and we 
denote with  $T(\logpliken) \in \bigO(\likecost)$ the cost of a single 
evaluation of the conditional likelihood. 

\subsection{Overall Complexity}
While the detailed derivations of 
the computational complexity are in  \apref{app:complexity}, 
here we state the main results.
Assuming that
$\k \ll \m$,  the total computational cost is given by:
\begin{align}
    T(\elbo) &\in\bigO(\q(\m^2 \d + \bsize\m\d + \k^2\m^3 +  \k\bsize\m^2 + \k\bsize\s\likecost))\text{,}\\
    \nonumber& \text{ and for diagonal posterior covariances we have:}\\
    T(\elbo) &\in \bigO(\q(\m^2 \d + \bsize\m\d + \m^3 + \k\bsize\m + \k\bsize\s\likecost))\text{.}
\end{align}
We note that it takes $\lceil \n/\bsize \rceil$ gradient updates to go over an entire pass of the data. If we assume that
$\k$ is a small constant, the asymptotic complexity of a single pass over the data does not improve by increasing
$\bsize$ beyond $\bsize=\m$. Hence if we assume that $\bsize \in \bigO(\m)$ the computational cost of a single gradient update is given by
\begin{align}
    T(\elbo) &\in\bigO(\q\m(\m\d + \m^2 + \s\likecost))\text{,}
\end{align}
for both diagonal and full covariances. If we assume that it takes a constant amount of time to compute the likelihood
function between a sample and an output, we see that setting $\s \in \bigO(\m\d + \m^2)$ does not increase the computational
complexity of our algorithm. As we shall see in section \ref{sec:med_expr}, even a complex likelihood function only requires $10,000$
samples to approximate it. Since we expect to have more than $100$ inducing points in most large scale
settings, it is reasonable to assume that the overhead from generated samples will not be significant in most cases.

\section{Experiments \label{sec:expts}}
In this section we analyze the behavior of our model on a wide range of experiments involving small-scale to large-scale datasets. The main aim of the experiments is to evaluate the performance of the model when considering different likelihoods and different dataset sizes.  We analyze how our algorithm's performance  is affected by the density and location of the inducing variables, and how the performance of batch and stochastic optimization compare. 

We start by evaluating our algorithm using five small-scale datasets ($N<1,300$) under different likelihood models and number  of inducing variables (\secref{sec:small_expr}). Then, using medium-scale experiments ($N<70,000$), we compare stochastic and batch optimization settings, and determine the effect of learning the inducing inputs on the performance (\secref{sec:med_expr}). Subsequently, we use \savigp on two large-scale datasets. The first one involves the prediction of airline delays, where we compare the convergence properties of our algorithm to models that leverage full knowledge of the likelihood  (\secref{sec:expts-airline}). The second large dataset considers an augmented version of the popular \mnist dataset for handwritten digit recognition, involving more than 8 million observations (\secref{sec:expts-mnist-large}). Finally, we showcase our algorithm on a non-standard inference problem concerning a seismic inversion task, where we show that our variational inference algorithm can yield solutions that closely match (non-scalable) sampling approaches (\secref{sec:seismic}).  Before  proceeding with the experimental set-up, we give details of our implementation which uses \gpu{s}. 

\subsection{Implementation}
We have implemented our \savigp method in Python and all the code 
is publicly available at \url{https://github.com/Karl-Krauth/Sparse-GP}.   
%
Most current mainstream implementations of Gaussian process models do not support \gpu computation, and instead opt to offload most of the work to the \cpu, with the notable exception of \cite{GPflow2017}. For example, neither of the popular packages 
\gpml\footnote{Available at \url{http://www.gaussianprocess.org/gpml/code/matlab/doc/}.}
or \gpy\footnote{Available at \url{https://github.com/SheffieldML/GPy}.} provide support for \gpu{s}.
This is despite the fact that matrix manipulation operations, which are easily parallelizable, 
are at the core of any Gaussian process model. 
In fact, the rate of progress between subsequent \gpu models has been much larger
than for \cpu{s}, thus ensuring that any \gpu implementation would run 
at an accelerated rate 
as faster hardware gets released. 

With these advantages in mind, we 
provide an implementation of  \savigp that uses Theano \citep{rfou2016theano}, 
a library that allows users to define symbolic mathematical
expressions that get compiled to highly optimized \name{gpu cuda} code. Any operation that
involved the manipulation of large matrices or vectors was done in Theano. Most of our experiments 
were either run on g2.2 \name{aws} instances, or on a desktop machine with an 
Intel core i5-4460 \cpu, 8GB of \name{ram}, and a \name{gtx760 gpu}.

Despite using a low-end  outdated  \gpu, 
we found  a time speed-up of  $5$x on average when we offloaded work to the \gpu. 
For example,
in the case of the \mnistbin dataset (used in section \ref{sec:expts-medium}), 
we averaged the time it took to compute ten 
gradient evaluations of the $\elbo$ with respect to the posterior parameters over the entire training set, 
where we expressed the posterior as a full Gaussian and used a sparsity factor of $0.04$.
While it took $42.35$ seconds, on average, per gradient computation when making use of the 
\cpu only, it took a mere $8.52$ seconds when work was offloaded to the \gpu. 
We expect the difference to be even greater given a high-end current-generation \gpu. 
\subsection{Details of the Experiments}
\paragraph{Performance measures.} We evaluated the performance of the algorithm using non-probabilistic 
and probabilistic measures according to the type of learning problem we are addressing.
The standardized squared error (\sse) and the negative log predictive density (\nlpd) were used in
the case of continuous-output problems. The error rate (\er) and the negative log probability (\nlp) were used in the
case of discrete-output problems. In the experiments using the airline dataset, we used the root mean squared error (\rmse)
instead of the \sse to be able to compare our method with previous work that used \rmse for performance evaluation.

\paragraph{Experimental settings.} In small-scale experiments, inducing inputs
were placed on a subset of the training data in a nested fashion, so that experiments on less sparse models contained the inducing points of the sparser models. In medium-scale and large-scale experiments
the location of the inducing points was initialized using the $k$-means clustering method. In all experiments the squared exponential covariance function was used.

\paragraph{Optimization methods.}
Learning the model involves optimizing 
variational parameters, hyperparameters, likelihood parameters, and inducing inputs.  
These were optimized iteratively in a global loop. In every iteration, each set of parameters was optimized separately while keeping the  other sets of parameters fixed, 
and this process was continued until the change in the objective function between two successive iterations was less than $10^{-6}$. For optimization in the batch settings,
each set of parameters was optimized using \name{l-bfgs}, with the maximum number of global iterations limited to $200$. In the case of stochastic optimization, we used the \adadelta method \citep{zeiler2012adadelta} with parameters $\epsilon = 10^{-6}$ and a decay rate of $0.95$. The choice of this optimization algorithm was motivated by (and to be consistent with) the work of \citet{hensmangaussian}, who found this specific algorithm successful in the context of Gaussian process regression. We compare with the method of \citet{hensmangaussian} in this context in \secref{sec:larg_expr}.

\paragraph{Reading the graphs.} Results are presented using boxplots and bar/line charts. In the case of boxplots, the lower, middle, and upper hinges correspond to the $25^{th}$, $50^{th}$, and $75^{th}$ percentile of the data. The upper whisker extends to the highest value that is within 1.5$\times$IQR of the top of the box, and the lower whisker extends to the lowest value within 1.5$\times$IQR of the bottom of the box (IQR is the distance between the first and third quantile). In the case of bar charts, the error bars represent $95\%$ confidence intervals computed over multiple partitions of the dataset. 

\paragraph{Model configurations.} We refer to the ratio of inducing points
to training points as the sparsity factor ($\sfmath = \m/\n$). For each sparsity factor, three different variations of \savigp corresponding to (i) a full Gaussian posterior, (ii) a diagonal Gaussian posterior, and  (iii) a mixture of two diagonal Gaussian posteriors were tested, which are denoted respectively by \full, \mix, and \mixtwo in the graphs.

\subsection{Small-scale Experiments}\label{sec:small_expr}
\begin{table}[t]
\begin{center}
\caption{Details of the datasets used in the small-scale experiments and their corresponding
 likelihood models.
 $\n_{train}, \n_{test}, \d$ are the number of training points, test points and input dimensions respectively;
 `likelihood' is the conditional likelihood used on each problem; and `model' is the name of 
 the model associated with that likelihood.  
 For \mining $\n_{test} = 0$ as only posterior inference is done (predictions on a test 
 set are not made for this problem).
 }
\label{table:datasets}
\begin{tabular}{r l l l l l}
\toprule
Dataset & \textbf{$\n_{train}$} & \textbf{$\n_{test}$} & \textbf{$\d$} & Likelihood $p(y|f)$ & Model   \\ 
\midrule
\mining & 811 & 0 & 1 & $\lambda^y \exp(-\lambda)/y!$ & Log Gaussian Cox process \\
\boston & 300 & 206 & 13 &$\Normal(y; f, \sigma^2) $ & Standard regression \\
\creep & 800 & 1266 & 30 & $\nabla_y t(y) \Normal(t(y); f, \sigma^2) $ & Warped Gaussian processes \\
\abalone & 1000 & 3177 & 8 & $\nabla_y t(y) \Normal(t(y); f, \sigma^2) $ & Warped Gaussian processes \\
\cancer & 300 & 383 & 9 &$1/(1+\exp(-f))$ & Binary classification \\
\usps & 1233 & 1232 & 256& $\exp(f_c) / \sum_{i=1} \exp(f_i)$ & Multi-class classification \\
\bottomrule
\end{tabular}
\end{center}
\end{table}

We tested \savigp on six small-scale datasets with different likelihood models. The datasets are summarized
in Table~\ref{table:datasets}, and are the same as those used by \citet{nguyen-bonilla-nips-2014}.
For each dataset, the model was tested five times across different subsets of the data; 
except for the mining dataset where only  training data were used for evaluation of 
the posterior distribution. 
 
\subsubsection{Standard regression}
\begin{figure}[t]
    \centering
    \begin{tabular}{c}
        \includegraphics[scale=1]{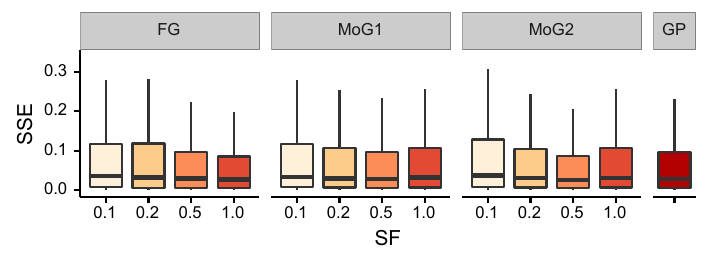} \\
        \includegraphics[scale=1]{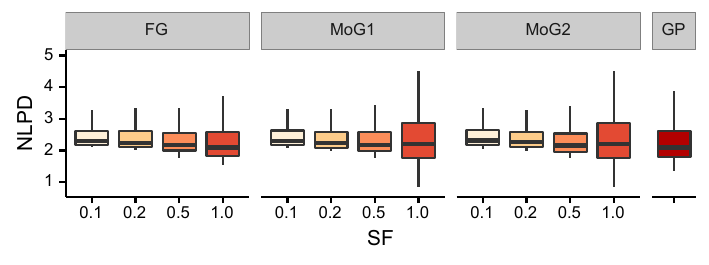}         
    \end{tabular}
    \caption{%
    The distributions of \sse and \nlpd for a regression problem with a univariate Gaussian likelihood model on the \boston housing dataset.
    Three approximate posteriors in \savigp are used:
     \full (full Gaussian),
     \mix (diagonal Gaussian), and 
     \mixtwo (mixture of two diagonal Gaussians), along with
      various sparsity factors ($\sfmath = M/N$). 
    The smaller the \sftext the sparser the model, with $\sfmath=1$ corresponding 
    to the dense model. \gptext corresponds to the performance of  exact inference  using   
    standard Gaussian process regression.
    }
    \label{fig:boston}%
\end{figure}

The model was evaluated on the \boston dataset \citep{uci2013}, which involves a standard regression problem with a univariate Gaussian likelihood model, i.e., $p(y_n|f_n)=\Normal(y_n|f_n,\sigma^2)$. Figure~\ref{fig:boston} shows the performance of \savigp for different sparsity factors, as well as the performance of exact Gaussian process inference (\gptext). As we can see, \sse increases slightly on sparser models. However, the \sse of all the models (\full, \mix, \mixtwo) across all sparsity factors are comparable to the performance of exact inference (\gptext). In terms of \nlpd, as expected, the dense ($\sfmath=1$) \full model  performs exactly 
like the exact inference method (\gptext). In the sparse models, \nlpd shows less variation in lower sparsity factors (especially for \mix and \mixtwo), which can be attributed  to the tendency of such models to make less confident predictions under high sparsity settings.
 
\subsubsection{Warped Gaussian process}
\begin{figure}[t]
    \centering
    \begin{tabular}{c}
        \includegraphics[scale=1]{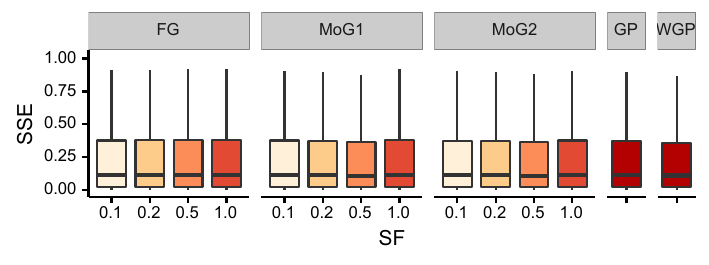} \\
        \includegraphics[scale=1]{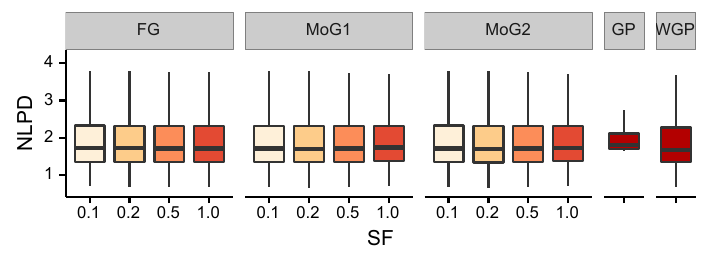} 
    \end{tabular}
    \caption{%
    The distributions of \sse and \nlpd for a warped Gaussian process likelihood model on the \abalone dataset.
    Three approximate posteriors in \savigp are used:
     \full (full Gaussian),
     \mix (diagonal Gaussian), and 
     \mixtwo (mixture of two diagonal Gaussians), along with
      various sparsity factors ($\sfmath = M/N$). 
    The smaller the \sftext the sparser the model, with $\sfmath=1$ corresponding 
    to the non-sparse model. \wgp corresponds to the performance of the exact inference method for warped Gaussian process models \citep{snelson2003warped}, and \gptext is the performance of  exact inference on a univariate Gaussian likelihood model.
    }
    \label{fig:abalone}%
\end{figure}

In warped Gaussian processes, the likelihood function is $p(y_n|f_n)=\nabla_y t(y_n) \Normal(t(y_n); f_n, \sigma^2)$, for some transformation $t$. We used the same neural-net style transformation as \citet{snelson2003warped}, and evaluated the performance of our model on two datasets: \creep \citep{cole2000modelling}, and \abalone \citep{uci2013}. The results are compared with
the performance of  exact inference  for warped Gaussian processes (\wgp) described by 
\citet{snelson2003warped}, and also with the performance of exact Gaussian process inference with a univariate Gaussian likelihood model (\gptext). As shown in Figure~\ref{fig:abalone}, in the case of the \abalone dataset, the performance is similar across all the models (\full, \mix and \mixtwo) and sparsity factors, and is comparable to the performance of the exact inference method for warped Gaussian processes (\wgp). The results on \creep are given in Appendix \ref{sec:smal-expts-creep}. 

\subsubsection{Binary classification}
\begin{figure}[t]
    \centering
    \begin{tabular}{c}
    \includegraphics[scale=1]{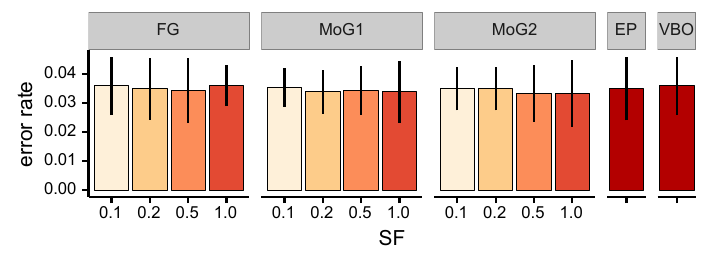} \\
    \includegraphics[scale=1]{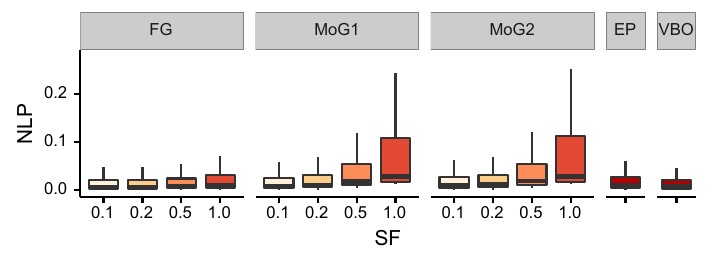}     
    \end{tabular}
    \caption{Error rates and 
    \nlp for binary classification using a logistic likelihood model on the 
     Wisconsin breast \cancer dataset. 
         Three approximate posteriors are used:
     \full (full Gaussian),
     \mix (diagonal Gaussian), and 
     \mixtwo (mixture of two diagonal Gaussians), along with
      various sparsity factors ($\sfmath = M/N$). 
    The smaller the \sftext the sparser the model, with $\sfmath=1$ corresponding to 
    the original model without sparsity. %
    The performance of inference using expectation propagation and
    variational bounds are denoted by \ep and \vbo respectively.
     }
    \label{fig:breast-cancer}
\end{figure}

For binary classification we used the logistic likelihood $p(y_n=1|f_n)=1/(1+e^{-f_n})$ on the breast \cancer dataset 
\citep{uci2013} and compared our model against the expectation propagation (\ep) and
variational bounds (\vbo) methods described by \citet{nickisch2008approximations}. Results depicted in
Figure~\ref{fig:breast-cancer} indicate that the error rate remains comparable across all models (\full, \mix, \mixtwo) and
sparsity factors, and is almost identical to the error rates obtained by inference using \ep and \vbo. Interestingly, the \nlp shows more variation and generally degrades as the number of inducing points is
increased, especially for \mix and \mixtwo models, which can be attributed to the 
fact that  these denser models are overly confident in their predictions.

\subsubsection{Multi-class classification}
\begin{figure}[t]
    \centering
\begin{tabular}{c}
    \includegraphics[scale=1]{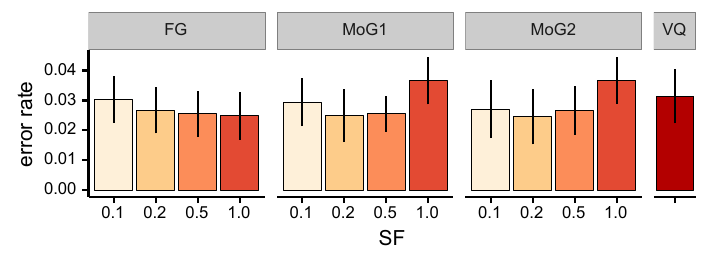} \\
    \includegraphics[scale=1]{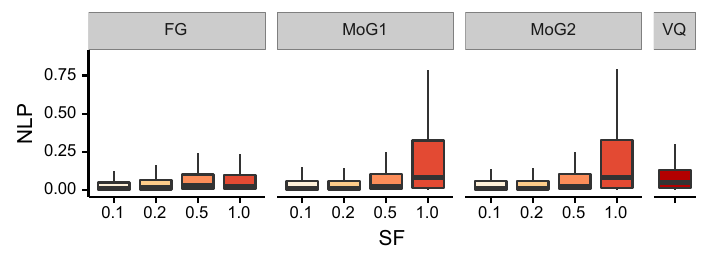}     
    \end{tabular}
    \caption{Classification error rates and
    \nlp for the 
     multi-class classification problem using a softmax likelihood model on the \usps dataset. 
         Three approximate posteriors in \savigp are used:
     \full (full Gaussian),
     \mix (diagonal Gaussian), and 
     \mixtwo (mixture of two diagonal Gaussians), along with
      various sparsity factors ($\sfmath = M/N$). 
    The smaller the \sftext the sparser the model, with $\sfmath=1$ corresponding to 
    the original model without sparsity. \vq corresponds to a variational inference method, which represents the \elbotext as a quadratic lower bound to the likelihood terms.
     }
    \label{fig:usps}
\end{figure}

For multi-class classification, we used the softmax likelihood $p(y_n=c)=e^{-f_c}/\sum_i e^{-f_i}$, and trained the model to classify the digits 4, 7, and 9 from the \usps dataset \citep{rasmussen-williams-book}. We compared our model against a variational inference method (\vq) which represents the \elbotext using a quadratic lower bound 
on the likelihood terms \citep{khan2012stick}. As we see in Figure~\ref{fig:usps}, the error rates are slightly lower in denser \full models. We also note that all versions of \savigp achieve comparable error rates to \vq's. 
Similarly to the binary classification case, \nlp shows higher variation with higher sparsity factor, especially in \mix and \mixtwo models.

\subsubsection{Log Gaussian Cox Process \label{sec:expts-lgcp}}
\begin{figure}[t]
    \centering
\begin{tabular}{cc}
    \includegraphics[scale=1]{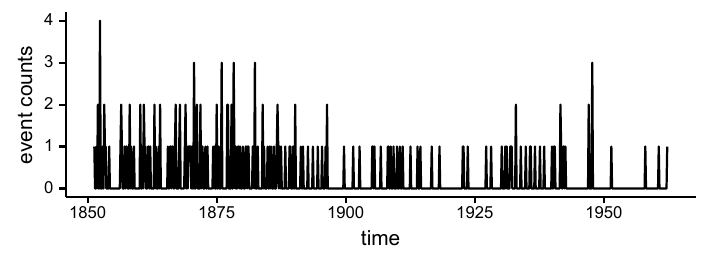} \\
    \includegraphics[scale=1]{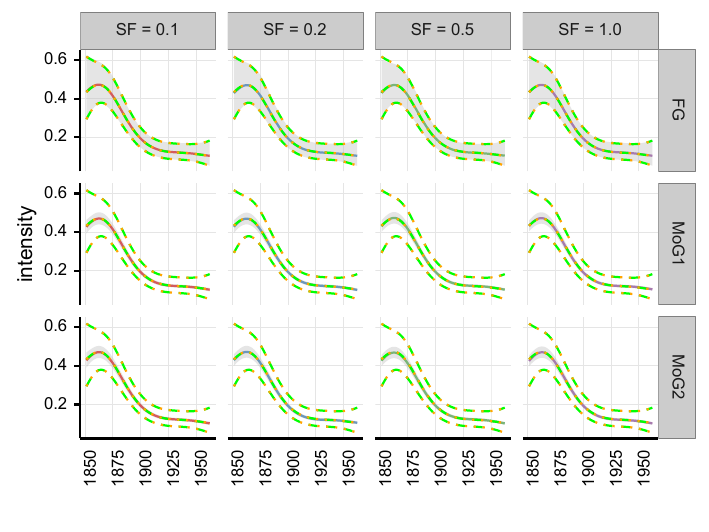}     
    \end{tabular}
    \caption{Top: the coal-mining disasters data. Bottom: the posteriors for a
    Log Gaussian Cox process on the data when using a
    \full (full Gaussian),
     \mix (diagonal Gaussian), and 
     \mixtwo (mixture of two diagonal Gaussians), along with various sparsity factors ($\sfmath = M/N$).
     The  smaller the \sftext the  sparser  the  model,  with  $\sfmath=1$  corresponding  to  non-sparse model. The solid line is the posterior mean and the shaded area represents 95\% confidence intervals. \hmc and \ess correspond to hybrid Monte Carlo and elliptical slice sampling inference methods and are represented by orange dots and dashed green lines respectively.
     }
    \label{fig:mining}
\end{figure}

The log Gaussian Cox process (\lgcp) is an inhomogeneous Poisson process in which the log-intensity
function is a shifted draw from a Gaussian process. Following \citet{murray2009elliptical},
we used the likelihood $\pcond{y_n}{f_n} = \frac{\lambda^{y_n}_{n}\exp{(-\lambda_n)}}{y_n!}$,
where $\lambda_n = \exp{f_n + m}$ is the mean of a Poisson distribution and $m$ is the offset
of the log mean. We applied \savigp with the \lgcp likelihood on a coal-mining disaster dataset \citep{jarrett1979note}, which can be seen in   
Figure \ref{fig:mining} (top). 

As baseline comparisons, we use hybrid Monte Carlo (\hmc) and elliptical slice sampling (\ess) described by \citet{duane1987hybrid} and
\citet{murray2009elliptical} respectively. We collected every $100^{th}$ sample for a total of $10k$
samples after a burn-in period of $5k$ samples and used
 the Gelman-Rubin potential scale reduction factors \citep{gelman1992inference} to check for convergence. 

The bottom plot of Figure~\ref{fig:mining} shows the mean and variance of the predictions made by \savigp, \hmc and \ess. We see that the \full model provides similar results across all sparsity factors, which is also comparable to the results provided by \hmc and \ess. \mix and \mixtwo models provide the same mean as the \full models, but tend to underestimate the posterior variance. This under-estimation of posterior variance is well known for variational methods, especially under factorized posteriors.
This results are more significant when comparing the running times across models. 
When using a slower Matlab implementation of our model, for a fair comparison  across all methods, 
\savigp was at least two orders of magnitude faster than \hmc and one order of magnitude faster than
\ess. 

\subsection{Medium-scale Experiments \label{sec:expts-medium}}\label{sec:med_expr}
In this section we investigate the performance of \savigp on
 four medium-scale problems on the datasets summarized in Table~\ref{tab:med_datasets}. 
Our goal here is to evaluate the model on medium-scale datasets, to study the effect of optimizing the inducing inputs, and to compare the performance of batch and stochastic optimization of the parameters. The first problem that we consider is multi-class classification of handwriting digits on the \mnist dataset using a softmax likelihood model. The second problem uses the same dataset, but the task involves binary classification of odd and even digits using the logistic likelihood model.
Therefore we refer to this dataset as the \mnist binary dataset (\mnistbin). The third problem uses the \sarcos dataset \citep{vijayakumar2000locally} and concerns 
an inverse-dynamics problem for a 
seven degrees-of-freedom \sarcos anthropomorphic robot arm. 
The task is to map from a 21-dimensional input space
 (7 joint positions, 7 joint velocities, 7 joint accelerations) to the corresponding 7 joint torques.
For this multi-output regression
problem we use the Gaussian process regression
network (\gprn) likelihood model of \citet{wilson-et-al-icml-12}, which allows for nonlinear models where the
correlation between the outputs can be spatially adaptive. This is achieved by taking a linear combination
of latent Gaussian processes, where the weights are also drawn from Gaussian processes. 
Finally, \sarcostwo is the same as \sarcos, but the model is learned using  only data from joints $4$ and $7$. 
For both \sarcos and \sarcostwo, the mean of the \sse and \nlpd across joints $4$ and $7$ are used for performance evaluation. We only consider joints $4$ and $7$ for \sarcos, despite the fact that
predictions are made across all $7$ joints to provide a direct comparison with \sarcostwo
and with previous literature \citep{nguyen-bonilla-uai-2014}.
We also made use of automatic relevance determination (\ard) for both the \sarcos and \sarcostwo datasets.

\begin{table}[t]
\begin{center}
\caption{Datasets used on the medium-scale experiments. \gprn stands for 
Gaussian process regression networks; 
$\n_{train}, \n_{test}, \d$ are the number of training points, test points and input dimensions respectively; 
$\q$ is the number of latent processes; 
$\p$ is the dimensionality of the output data;
and `model' is the model associated with the conditional likelihood used. }
\label{tab:med_datasets}
\begin{tabular}{l l l l l l l }
\toprule
Dataset    & \textbf{$N_{train}$}	 & \textbf{$N_{test}$} 	& \textbf{$D$} 	& $Q$ 	& $P$ 	& Model   \\ 
\midrule 
\mnistbin & 60,000 			 	& 10,000 				&  784 			& 1 	& 1 	& Binary classification \\
\mnist 	   &  60,000  				& 10,000 				&  784 			& 10 	& 10 	& Multi-class classification \\
\sarcostwo &  44,484  			&  4,449 				& 21 			& 3 	& 2 	& \gprn \citep{wilson-et-al-icml-12} \\
\sarcos &  44,484  				&  4,449 				& 21 			& 8 	& 7 	& \gprn \citep{wilson-et-al-icml-12} \\
\bottomrule
\end{tabular}
\end{center}
\end{table}

\subsubsection{Batch optimization}
\paragraph{Classification.} 
Here we evaluate the performance of batch optimization of model parameters on multi-class classification using \mnist  and refer the reader to  Appendix \ref{sec:medium-binary-mnistbin} for the  results on binary classification using \mnistbin. We optimized the kernel hyperparameters and the variational parameters, but fixed the inducing point locations using $k$-means clustering. Unlike most previous approaches \citep{ranzato2006energy, jarett2009recognition}, we did not tune model parameters using the validation set. Instead, we considered the validation set to be part of the training set and used our variational framework to learn all model parameters. As such, the current setting likely provides a lower performance on test accuracy compared to the approaches that use a validation dataset, however our goal is simply to show that we were able to achieve competitive performance in a sparse setting when no knowledge of the likelihood model is used.

\begin{figure}[t]
	\centering
	\begin{tabular}{cc}
		\includegraphics[scale=1]{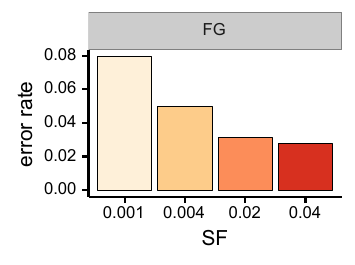} &
		\includegraphics[scale=1]{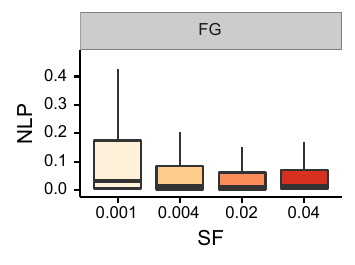} 		
	\end{tabular}
	\caption{Error rates and 
	\nlp for multi-class classification on the 
	 \mnist dataset. 
	 We used a full Gaussian (\full) posterior approximation across
	  various sparsity factors ($\sfmath = M/N$). 
	The smaller the SF the sparser the model.
	 }
	\label{fig:mnist_full}%
\end{figure}

Figure~\ref{fig:mnist_full} shows the result on the \mnist dataset. 
we see that the performance improves with denser models.
Overall, \savigp achieves an error rate of $2.77\%$ at $\sfmath = 0.04$.
This is a significant improvement over the results reported by \citet{gal-et-al-nips-2014}, in which a separate model was trained for each digit, which achieved an error rate of $5.95\%$. As a reference, previous literature
reports about $12\%$ error rate by linear classifiers and less than $1\%$ error rate by state-of-the-art
deep convolutional nets. Our results show that our method reduces the gap between \gptext{}s and deep nets
while solving the harder problem of full posterior estimation. In \apref{sec:medium-binary-mnistbin} we show that our model can achieve slightly better performance than that reported by \cite{hensman-et-al-aistats-2015} on \mnistbin.

\begin{figure}[t]
	\centering
	\begin{tabular}{cc}
		\includegraphics[]{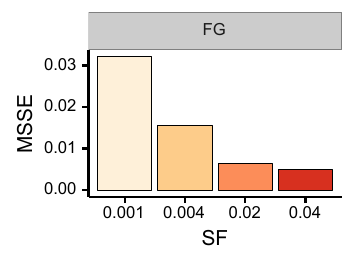} &
		\includegraphics[]{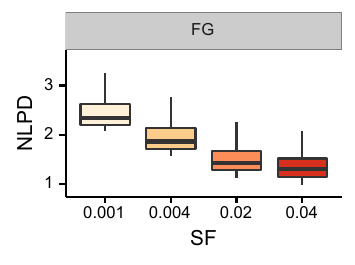} 		
	\end{tabular}
	\caption{Mean \sse and 
	\nlpd for multi-output regression on the 
	 \sarcostwo dataset. 
	 We used a full Gaussian (\full) posterior approximation across
	  various sparsity factors ($\sfmath = M/N$). 
	The smaller the SF the sparser the model.
	 }
	\label{fig:sarcos_two}%
\end{figure}

\paragraph{Gaussian process regression networks.} 
Figure~\ref{fig:sarcos_two} shows
that \sarcostwo gets a significant benefit from a higher number of inducing points,
which is consistent with previous work that found that the performance on this dataset improves as more data is being used to train the model \citep{vijayakumar2000locally}. The performance is significantly better than the results reported by \citet{nguyen-bonilla-uai-2014},
who achieved a mean standardized squared eror (\msse) of $0.2631$ and $0.0127$ across joints $4$ and $7$, against our
values of $0.0033$ and $0.0065$ for $\sfmath = 0.04$. However, we note that their setting was much sparser than ours on joint $4$.
The results on \sarcos (predicting on all joints) are given in  \apref{sec:expts-medium-gprn-sarcos}.

\subsection{Inducing-input learning \label{sec:expts-learn-z}}
We now compare the effect of adaptively learning the inducing inputs,
versus initializing them using $k$-means and leaving them fixed. We look at the performance
of our model under two settings: (i) a low number of inducing variables (SF$=0.001, 0.004$) where the
inducing inputs are learned,
and (ii) a large number of inducing variables ($\sfmath=0.02, 0.04$) without learning of 
their locations.

\begin{figure}[t]
	\centering
	\begin{tabular}{cc}
		\includegraphics[]{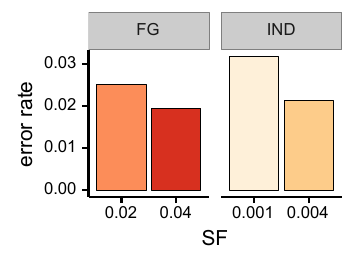} &
		\includegraphics[]{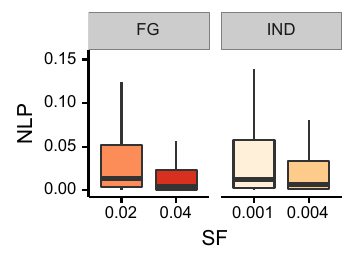} 		
	\end{tabular}
	\caption{Comparison of error rate and 
	\nlp obtained by \savigp without learning (\full) and with learning (\ind)
	 of inducing inputs for binary classification on the 
	 \mnistbin dataset.}
	\label{fig:mnist_bin_ind}%
\end{figure}

Figure~\ref{fig:mnist_bin_ind} shows the performance of the model under the two settings on \mnistbin. We see that learning the location of the inducing variables yields a large gain in performance. In fact, the sparser models with inducing point learning performed similarly to the denser models, despite the fact that the two models differed by an order of magnitude when it came to the number of inducing variables. Additional results on \mnist, \sarcos and \sarcostwo are shown in Appendix \ref{sec:expts-medium-inducing-mnist-sarcos}. Our analyses indicate that  there is a trade-off between the reduction in computational complexity gained by reducing the number of inducing variables and the increased computational cost of calculating inducing-input gradients. 
As such, the advantage of learning the inducing inputs is dataset dependent and it is affected mainly by 
the input dimensionality ($\d$).

\subsection{Batch Optimization vs Stochastic Optimization}
In this section we compare the performance of our model after it has been trained in a
batch setting versus a stochastic setting. We used \adadelta as our stochastic
optimization algorithm as it requires less hand-tuning than other algorithms such as
vanilla stochastic gradient descend (\sgd). 

\begin{table}[t]
\begin{center}
\caption{Error rate (\er) and negative log probability (\nlp) obtained with \savigp 
	optimized using  batch optimization (with \name{l-bfgs}) and 
	stochastic optimization (with \adadelta) on the \mnistbin dataset.
         The inducing inputs are optimized in both cases.}
\label{tab:batch_stoch}
\begin{tabular}{l p{1.9cm} p{1.9cm} p{1.9cm} p{1.9cm}  }
\toprule
Method    & Batch $\sfmath=0.001$ & Batch $\sfmath=0.004$ & Stochastic $\sfmath=0.001$ &  Stochastic $\sfmath=0.004$ \\ 
\midrule 
\er       & 3.17\%        & 2.12\%   & 3.11\%          & 2.68\%\\
\nlp      & 0.097  	      & 0.068     & 0.099           & 0.083 \\
\bottomrule
\end{tabular}
\end{center}
\end{table}
Table~\ref{tab:batch_stoch} shows only a slight deterioration in predictive performance on \mnistbin when using stochastic optimization instead of batch optimization. In fact, our exploratory experiments showed that the 
error metrics could have significantly been reduced  by meticulously hand-tuning momentum 
stochastic gradient descent (\name{sgd}). However, our goal was simply to show that our model does not suffer from a large loss in performance when going from a batch to a stochastic setting in medium-scale experiments,
 without requiring extensive hand tuning of the optimization algorithm.
As we shall see in the next section, when batch optimization is not feasible, 
stochastic optimization in our model performs well when compared to state-of-the-art 
approaches for inference in \gptext models on very large datasets.

\subsection{Large-scale Experiments}
\label{sec:larg_expr}
In this section we evaluate \savigp  on two large-scale problems  involving prediction of airline delays (regression, $N=700,000$) and  handwritten digit recognition (classification, $N=8,100,000$). 
\subsubsection{Airline delays}\label{sec:expts-airline}
Here we consider the problem of predicting airline delays using a univariate Gaussian likelihood model \citep{hensmangaussian}. 
We note  that this dataset has outliers that can significantly affect the performance of 
regression algorithms when using metrics such as the \rmse. 
This has also been pointed out by \citet{das-et-al-arxiv-2015}.
Additionally, in order to match the application of the algorithms to a realistic setting, 
evaluation of learning algorithms on this dataset should always consider making predictions in the future. 
Therefore, unlike \cite{hensmangaussian}, 
we did not randomly select the training and test sets, instead we selected the first
$700,000$ data points starting at a given offset as the training set and the next $100,000$ data
points as the test set. We generated five training/test sets by setting the initial offset to $0$ and increasing it by $200,000$ each time.

We used the squared exponential covariance function with automatic relevance determination (\ard), and optimized all the parameters using \adadelta \citep{zeiler2012adadelta}. We compare the performance of our model with the stochastic variational
inference on \gptext{s}  method  (\svigp) described by \citet{hensmangaussian}, which assumes full knowledge of the likelihood function.
 Our method (\savigp) and \svigp  were optimized using \adadelta with identical settings.
 We also report the results of Bayesian linear regression with a
zero-mean unit-variance prior over the weights (\name{linear}) 
and  Gaussian process regression using  subsets of
the training data (\gpone and \gptwo using $1,000$ and $2,000$ datapoints respectively).
For each run we fit \gpone and \gptwo ten times using a randomly selected subset of the training data. Kernel hyper-parameters 
of \gpone and \gptwo were optimized using \name{l-bfgs}.

\begin{figure}[t]
	\centering
	\begin{tabular}{cc}
		\includegraphics[scale=1]{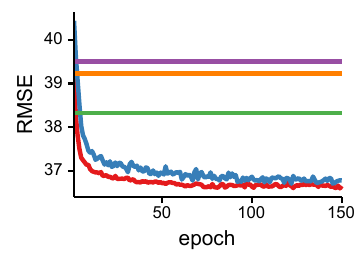} &
		\includegraphics[scale=1]{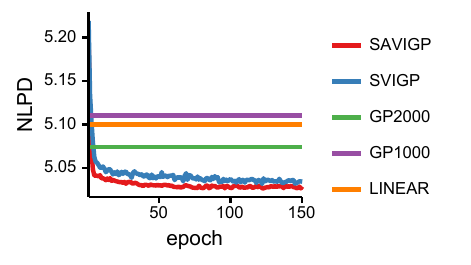} 
	\end{tabular}
	\caption{%
	The \rmse (left) and \nlpd (right) of \svigp \citep{hensmangaussian}, and 
	\savigp (our method) averaged across all $5$ airline-delay experiment runs.
	The x-axis represents the  number of passes over the data (epochs). The figure also shows the performance
	 of \gpone{}, \gptwo and Bayesian linear regression (\name{linear}) after training is completed.
	}
	\label{fig:airline_performance}%
\end{figure}

Figure~\ref{fig:airline_performance} shows the performance of \savigp and the four baselines.
We see that \savigp converges at a very similar rate to \svigp, despite making no assumption about the likelihood model. Furthermore, \savigp performs better
than all three simple baselines after less than $5$ epochs. We note that our results are not directly comparable
with those reported by \citet{hensmangaussian}, since we deal with the harder problem of predicting future
events due to the way our dataset is selected. 
\subsubsection{Training Loss vs Test Performance}
\begin{figure}[t]
	\centering
	\begin{tabular}{cc}
		\includegraphics[scale=1]{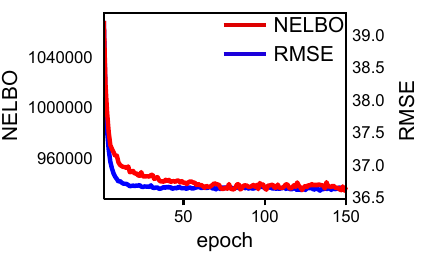}\quad &
		\includegraphics[scale=1]{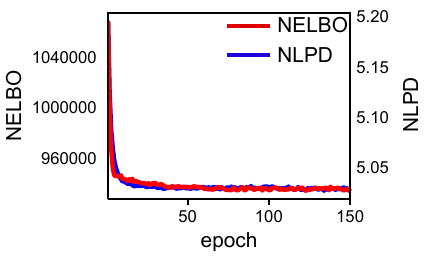} 
	\end{tabular}
	\caption{%
	The \rmse (left) and \nlpd (right) of \savigp on the test data
	alongside the negative evidence lower bound (\name{nelbo}) on the training data,
	averaged over five runs of the airline-delay experiment.
    }
	\label{fig:airline_elbo}%
\end{figure}

An additional interesting question about the behavior of our model relates to 
how well the training objective function correlates to the test error metrics. 
Figure~\ref{fig:airline_elbo} shows how the performance metrics (\rmse and \nlpd)
on the test data vary along with the negative evidence lower bound (\name{nelbo}).
 As we can see, changes in the \name{nelbo} closely mirror changes in 
 \rmse and \nlpd in our model, which is an indication that the variational objective 
 generalizes well to test performance when using these metrics.

\subsubsection{Large-scale mnist}\label{sec:expts-mnist-large}
In this section we evaluate the performance of \savigp on the \mnistlarge \citep{loosli-canu-bottou-2006} dataset, which artificially extends the \mnist dataset to 8.1 million training points by pseudo-randomly transforming existing \mnist images.

We train \savigp on the \mnistlarge dataset by optimizing only variational parameters stochastically, with a batch size of $1000$ and $2000$ inducing points. We also use a squared exponential covariance function without automatic relevance determination.

After optimizing the model for $19$ hours, we observe an error rate of $1.54\%$ on the test set and lower, middle and upper \nlp quartiles of $8.61\mathrm{e}{-4}$, $3.81\mathrm{e}{-3}$, and $2.16\mathrm{e}{-2}$ respectively. We see that this  outperforms significantly  our previous result in \secref{sec:expts-medium} on standard \mnist, where we reported an error rate of $2.77\%$.  As a point of comparison with hard-coded approaches,  \cite{henao2012predictive}  reported 0.86\% error rate  on \mnist  when  considering an augmented  active set (which is analogous to the concept of inducing inputs/variables) and a 9th-degree polynomial covariance function.

\begin{figure}[t]
    \centering
    \begin{tabular}{cc}
        \includegraphics[scale=1]{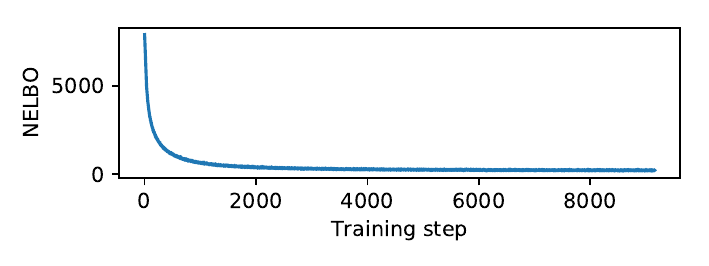}
    \end{tabular}
    \caption{%
        The negative evidence lower bound  (\name{nelbo}) on the current batch of the \mnistlarge training set over time on \savigp. The x-axis represents the number of training steps taken.
    }
    \label{fig:mnist8m_performance}%
\end{figure}

Finally, Figure~\ref{fig:mnist8m_performance} shows the training loss (the \name{nelbo}) of \savigp on \mnistlarge as a function of  the number of training steps, where we note that the loss decreases rapidly in the first 2,000 iterations and stabilizes at around 4,000 iterations. 

 \subsection{Seismic inversion} \label{sec:seismic}
\newcommand{\fdepth}{f^{\text{d}}}
\newcommand{\fvel}{f^{\text{v}}}
In this experiment we aim to evaluate our method qualitatively on a non-standard inference problem motivating the need for black-box likelihoods. This application considers a one-dimensional seismic inversion and was investigated by \citet{bonilla2016extended}. Given noisy surface observations of sound-reflection times, the goal is to infer the geometry (layer depths) of subsurface geological layers and seismic propagation velocities within each layer. For this, we consider a real dataset from a seismic survey carried out in the Otway basin region in the state of Victoria, Australia. 

\paragraph{Setting} There are $\n=113$ site locations with $\p=4$ interface reflections (layers) per site. The inputs are given by the surface locations of the seismic sensors ($\X$), the outputs are the observed times at the different locations for each of the layers ($\Y$) and we aim to infer $\q= 2*\p$ latent functions, i.e.~$\p$ functions corresponding to layer depths and $\p$ functions corresponding to seismic propagation velocities. For clarity, in this section we refer to the latent functions corresponding to depth and velocity as $\fdepth$ and $\fvel$, respectively. 

\paragraph{Likelihood} The likelihood of the observed times $y_{np}$ for location $\x_n$  and interface $p$  is given by:
\begin{equation}
	y_{np} =  \left\lbrace 
\begin{array}{ll}
	2 \left( \frac{\fdepth_{np}}{\fvel_{np}} \right) + \epsilon_{np}, & \text{ for } p=1,  \\
		2 \left( \frac{\fdepth_{np} - \fdepth_{np-1}}{\fvel_{np}} \right) + y_{np-1} + \epsilon_{np}, &  \text{ for } 1 < p \leq P \text{,}
\end{array}
\right.
\end{equation}
where  $\epsilon_{np} \sim \Normal(0, \sigma^2_{p})$ and $\sigma^2_{p}$ is the output-dependent noise variance.  As in \cite{bonilla2016extended}, we set the corresponding standard deviations to 0.025s, 0.05s, 0.075s and 0.1s. 

\paragraph{Prior setting:} 
We used the same prior as in \citet{bonilla2016extended}, with prior mean depths of 200m, 500m, 1600m and 2200m and prior mean velocities of 1950m/s, 2300m/s,
2750m/s and 3650m/s. The corresponding standard deviations for the depths were set to 15\% of the layer mean, and for the velocities they were set to 10\% of the layer mean. A squared exponential covariance function with unit length-scale was used. 
 
\begin{figure}
	\includegraphics[width=0.45\textwidth]{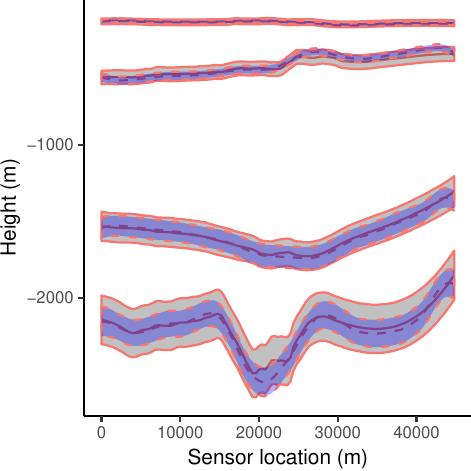} 
	\includegraphics[width=0.45\textwidth]{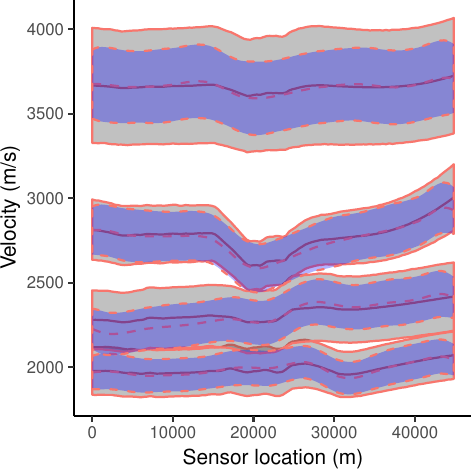} 
	\caption{Results for the seismic inversion experiment using our algorithm (\savigp) and the \mcmc algorithm developed by \citet{bonilla2016extended}. The mean and standard deviations envelopes are shown for \savigp and \mcmc in solid and dashed lines, respectively. Left: inferred layer boundaries. Right: inferred seismic velocities.  
	\label{fig:seismic}}
\end{figure}

\paragraph{Posterior estimation} We ran our algorithm for the dense case ($\Z = \X$) using a full Gaussian posterior and batch optimization, initializing the posterior means to the prior means and the posterior covariance to a diagonal matrix with entries corresponding to $0.1\%$ of the prior variances.  The results are given in Figure \ref{fig:seismic}, where we see that \savigp's posterior closely match the ``true" posterior obtained by the \mcmc algorithm developed in \citet{bonilla2016extended}, although the variances are overestimated, which can be see as a consequence of our variational approach using a full Gaussian approximate posterior.

\section{Conclusions and Discussion \label{sec:conclusion}}
We have developed 
\emph{scalable automated variational inference for Gaussian process} models (\savigp), 
an inference method for models with Gaussian process (\gptext) priors,
multiple outputs,  and nonlinear likelihoods. The method is generally applicable 
to black-box likelihoods, i.e.~it does not need to know the 
 details of the conditional likelihood (or its gradients), 
 only requiring its evaluation  as a black-box function.

 One of the key properties of this method is that, 
 despite using a flexible variational posterior such as a  
 mixture-of-Gaussians distribution,  it is 
 \emph{statistically efficient} in that it requires samples 
 from univariate Gaussian distributions to estimate the 
 evidence lower bound (\elbotext) and its gradients. 
  
  In order to provide scalability to very large datasets, 
  we have used an augmented prior via the so-called inducing variables, 
   which are prevalent in most sparse \gptext approximations. This has allowed 
   us to decompose the \elbotext as a sum of terms over the datapoints, 
   hence giving way for the application of stochastic optimization and 
   parallel computation. 
   
  Our small-scale experiments have shown that \savigp can perform as 
   accurately as solutions that have been hard-coded specifically for the 
   conditional likelihood of the problem at hand, even under high 
   sparsity levels. 

Our medium-scale experiments also have shown the effectiveness of \savigp, 
when considering problems such as multi-class classification on 
the \mnist dataset and highly nonlinear likelihoods such as that 
used in \gprn. On these experiments, we also have analyzed the effect of learning 
the inducing inputs, i.e.~the locations corresponding to the inducing variables, and 
concluded that doing this can yield significant improvements in performance, reducing 
the numbers of inducing variables required  to achieve similar accuracy by an 
order of magnitude. Nevertheless, there is generally a trade-off between reducing 
the time cost gained by having a lower number of inducing inputs and 
the overhead of carrying out optimization over these locations. 

 Our first large-scale experiment, on the problem of predicting airline delays,  
showed that \savigp is on par with the state-of-the-art approach for 
Gaussian process regression on big data \citep{hensmangaussian}. In fact, 
our results show that \savigp is slightly better but we attribute those differences 
to implementation specifics. The important message is that
\savigp can attain state-of-the-art performance even without exploiting specific 
knowledge of the likelihood model. Our second large-scale experiment shows that 
having an inference algorithm for \gptext models with non-Gaussian likelihood for large datasets is worthwhile, as
 the performance obtained on \mnistlarge (using 8.1M observations) was significantly better 
 than on the standard \mnist(using 60K observations). 

Our final experiment considered a non-standard inference problem concerning a seismic 
inversion task. In this problem \savigp yielded a solution for the posterior over latent 
functions that closely matched the solution obtained by a (non-scalable) sampling algorithm. 

Overall, we believe \savigp has the potential to be a powerful tool for  
 practitioners and researchers 
 when devising models for new or existing problems with 
 Gaussian process priors  for which variational inference 
is not yet available. 
As mentioned in Section \ref{sec:related},  
we are very much aware of recent developments in the areas 
of probabilistic programming, stochastic variational inference 
and Bayesian deep learning. Advances such as those  
in black-box variational inference \citep[\bbvi,][]{ranganath2014black}
and variational auto-encoders \citep{rezende2014stochastic,kingma2013auto} 
are incredibly exciting for the machine learning community. 
While the former, \bbvi, is somewhat too general to be useful 
in practice for \gptext models, the latter (variational auto-encoders) 
requires specific knowledge of the likelihood and is not 
a truly black-box method. 


Finally, we are  working on extending our models to 
more complex settings such as structured prediction problems, 
i.e.~where the conditional likelihood is a non-iid model such as 
a chain or a tree \citep[see e.g.][for a recent reference]{galliani-et-al-2017}. Such settings provide incredible challenges from 
the computational and modeling perspectives. For example, 
how to deal with the exponential increase in the number of parameters 
of the model, and how to reduce the number of calls to an 
expensive conditional likelihood model.
We believe that the benefits of being Bayesian 
well outweigh the effort in exploring such challenges.


\acks{}
We acknowledge the contribution by Trung V.~Nguyen to the original 
conference paper \citep{nguyen-bonilla-nips-2014}. 
\evb  started this work at the University of New South Wales (\name{unsw sydney}) and was partially  supported 
by
\name{unsw}'s Faculty of Engineering Research Grant Program
project \# PS37866;
\name{unsw}'s  Academic Start-Up Funding Scheme project \# PS41327; 
and an \name{aws} in Education Research Grant award.
 AD started this work at \name{unsw sydney}, and was supported by a Research Fellowship from \name{unsw sydney}.

\appendix
\section{Derivation of the KL-divergence Term \label{app:klterm}}
Here we derive the expressions for the terms composing the KL-divergence ($\klterm = \enterm + \crossterm$) 
part of the log-evidence lower bound ($\elbo$).

The entropy term is given by:
\begin{align}
\enterm(\llambda) &=  - \expectation{\qu}{\log \qu} \\
&= 
- \int \sum_{k=1}^\k \pi_k q_k(\u | \postmean{k}, \postcov{k})  
\log \sum_{\ell=1}^\k  \pi_\ell q_\ell(\u | \postmean{\ell}, \postcov{\ell})  \der \u \\
\label{eq:jensen1}
&=
- \sum_{k=1}^\k \pi_k 
\int \Normal(\u; \postmean{k}, \postcov{k})  \log \sum_{\ell=1}^\k  
\pi_\ell 
\Normal(\u ; \postmean{\ell}, \postcov{\ell})  \der \u \\
\label{eq:jensen2}
&  \geq 
- \sum_{k=1}^\k \pi_k \log \int  \Normal(\u ; \postmean{k}, \postcov{k}) 
  \sum_{\ell=1}^\k  \pi_\ell \Normal(\u | \postmean{\ell}, \postcov{\ell})  \der \u \\
&=
- \sum_{k=1}^\k \pi_k \log  \sum_{\ell=1}^\k  \pi_\ell \int  \Normal(\u ; \postmean{k}, \postcov{k})  \Normal(\u | \postmean{\ell}, \postcov{\ell})  \der \u \\
&= 
- \sum_{k=1}^\k \pi_k \log  \sum_{\ell=1}^\k  \pi_\ell   \Normal(\postmean{k} ;  \postmean{\ell}, \postcov{k} + \postcov{\ell}  ) 
	\defeq \entermhat \text{,}	
\end{align}
where we have used Jensen's inequality to bring the logarithm out of the integral from 
Equations \eqref{eq:jensen1} to \eqref{eq:jensen2}. 

The negative cross-entropy term can be computed as:
\begin{align}
\crossterm(\llambda)
&=
\expectation{\qu}{\log p(\u)}  
=  \sum_{k=1}^\k \int \pi_k \qku \log p(\u) \der \u \\
&= \sum_{k=1}^\k \sum_{j=1}^\q \pi_k \int \Normal(\u_j; \postmean{kj}, \postcov{kj}) \log \Normal(\u_j; \vec{0}, \Kzz) \der \u_j \\
&= \sum_{k=1}^\k \sum_{j=1}^\q \pi_k \left[ \log \Normal(\postmean{kj}; \vec{0}, \Kzz) - \frac{1}{2} \trace \Kzzinv \postcov{kj}
 \right] \\
&= - \frac{1}{2} \sum_{k=1}^\k \pi_k
\sum_{j=1}^\q \left[ \m \log 2 \pi  + \log \det{\Kzz} + \postmean{kj}^T \Kzzinv \postmean{kj} + \trace \Kzzinv \postcov{kj} \right]
\text{.}
\end{align}
%
\section{Proof of Theorem \ref{th:univariate} \label{app:proof-th-univariate}}
In this section we prove Theorem \ref{th:univariate} concerning the statistical 
efficiency of the estimator of the expected log likelihood and its gradients, i.e.~that both
can be estimated using expectations over Univariate Gaussian distributions.
\subsection{Estimation of $\ellterm$}  
Taking the original expression in Equation \eqref{eq:ellterm-original} we have that:  
\begin{align}
\label{eq:proof-ell1}
\ellterm(\llambda) &  =  \expectation{\qf}{\log \plike} \text{,} \\ 
\label{eq:proof-ell2}
&= \sum_{n=1}^{\n}  \int_{\f} {\qf} {\logpliken} \der \f  \text{,}\\
\label{eq:proof-ell3}
&= \sum_{n=1}^{\n} \int_{\fn} \int_{\fnotn} q(\fnotn | \fn) q(\fn) \logpliken  \der \fnotn \der \fn \text{,} \\
\label{eq:proof-ell4}
&= \sum_{n=1}^{\n} \expectation{\qnf}{\logpliken}  \text{,} \\
\label{eq:proof-ell5}
&= \sum_{n=1}^{\n} \sum_{k=1}^\k \pi_k \expectation{\qknf}{\logpliken}  \text{,}
\end{align}
where we have applied the linear property of the expectation operation in 
Equation \eqref{eq:proof-ell2}; 
used $\fnotn$ to denote all the latent functions except those corresponding to the n$\mth$ observation, 
and integrated these out to obtain  Equation  \eqref{eq:proof-ell4};
and used the form of the marginal posterior (\mog in Equation \eqref{eq:qf}) to get
 Equation \eqref{eq:proof-ell5}. 
 
 As described in Section \ref{sec:ell}, $\qknf$ is a $\q$-dimensional Gaussian with diagonal 
 covariance, hence computation of Equation \eqref{eq:proof-ell5} only requires expectations 
 over univariate Gaussian distributions.
\QEDA
\subsection{Estimation of the Gradients of $\ellterm$}
Denoting the k$\mth$ term for the n$\mth$ observation of the expected log likelihood with $\elltermkn$ we have that:
\begin{align}
  \elltermkn &=   \expectation{\qknf}{\logpliken} \\
       & = \int_{\fn}  \qknf   \logpliken \der \fn \\
       \label{eq:grad-full-1}
\grad_{\llambda_k}  \elltermkn &=    \int_{\fn} \qknf  \grad_{\llambda_k} \log \qknf \logpliken \der \fn \\
\label{eq:grad-full-2}
& = \expectation{\qknf}{\grad_{\llambda_k} \log \qknf \logpliken} \text{,}
\end{align}
for $\llambda_k \in \{\postmean{k}, \postcov{k} \}$, and for the 
mixture proportions the gradients can be estimated straightforwardly using Equation 
\eqref{eq:ellgradients-2}. 
We have used in Equation \eqref{eq:grad-full-1} the fact that $\grad_{\vec{x}} f(\vec{x}) = f(\vec{x}) \grad_{\vec{x}} \log f(\vec{x})$ for 
any nonnegative function 
$f(\vec{x})$. We see that our resulting gradient estimate has an analogous form to that obtained in Equation \eqref{eq:proof-ell5}, 
hence its computation only requires expectations  over univariate Gaussian distributions.
\QEDA
%
%
\section{Gradients of the Evidence Lower Bound wrt Variational Parameters \label{app:gradients-variational}}
Here we specify the gradients of the log-evidence lower bound ($\elbo$) wrt variational parameters. 
For the covariance, we consider $\postcov{kj}$ of general structure but also give the updates 
when $\postcov{kj}$ is a diagonal matrix, denoted with $\postcovdiag{kj}$. 
\subsection{KL-divergence Term}
Let $\Kzzall$ be the block-diagonal covariance with $\q$ blocks $\Kzz$, $j=1, \ldots \q$.  Additionally,
let us assume the following definitions:
\begin{align}
	\label{eq:Ckl}
	\Ckl  & \defeq \postcov{k} + \postcov{\ell} \text{,}\\
	\nkl  & \defeq \Normal(\postmean{k}; \postmean{\ell}, \Ckl) \text{,} \\
	z_k  &\defeq \sum_{\ell=1}^\k \pi_{\ell} \nkl \text{.}
\end{align}
The gradients of 
 $\klterm$ wrt the posterior mean and posterior covariance for component $k$ are: 
\begin{align}
\grad_{\postmean{k}} \crossterm &= - \pi_k \Kzzallinv \postmean{k}  \text{,}\\
\grad_{\postcov{k}} \crossterm &= - \frac{1}{2} \pi_k \Kzzallinv  \text{, and for diagonal covariance we have:}\\
\grad_{\postcovdiag{k}} \crossterm &= - \frac{1}{2} \pi_k \diag (\Kzzallinv)  \text{,} \\\
\grad_{\pi_k} \crossterm &= - \frac{1}{2}  \sum_{j=1}^\q [\m \log 2 \pi + \log \det{\Kzz} 
	+ \postmean{kj}^T \Kzzinv \postmean{kj} + \trace{\Kzzinv \postcov{kj}}]	
\text{,}
\end{align}
where we note that  we compute $\Kzzallinv$ by inverting the corresponding blocks $\Kzz$ independently. 
The gradients of the entropy term wrt the variational parameters are:
\begin{align}
	\label{eq:grad-ent-init}
	\grad_{\postmean{k}} \entermhat &= \pi_k \sum_{\ell=1}^\k \pi_{\ell} \left( \frac{\nkl}{\zk} + \frac{\nkl}{\zl} \right)   \Ckl^{-1} (\postmean{k} - \postmean{\ell}) \text{,} \\
	\grad_{\postcov{k}} \entermhat &= \frac{1}{2} \pi_k \sum_{\ell=1}^{\k} \pil \left(  \frac{\nkl}{\zk} + \frac{\nkl}{\zl} \right) 
	\left[ \Ckl^{-1} - \Ckl^{-1}  (\postmean{k} - \postmean{\ell})  (\postmean{k} - \postmean{\ell})^T \Ckl^{-1}  \right] \text{,} \\
	\nonumber
	& \text{ and for diagonal covariance we have:} \\
	\grad_{\postcovdiag{k}} \entermhat &= \frac{1}{2} \pi_k \sum_{\ell=1}^{\k} \pil \left(  \frac{\nkl}{\zk} + \frac{\nkl}{\zl} \right) 
	\left[ \tildeCkl^{-1} - \tildeCkl^{-1} \diag \left(  (\postmean{k} - \postmean{\ell}) \hada (\postmean{k} - \postmean{\ell}) \right) \tildeCkl^{-1}  \right]  \text{,}\\
	\label{eq:grad-ent-end}
	\grad_{\pi_k} \entermhat &= - \log \zk - \sum_{\ell=1}^{\k} \pi_{\ell} \frac{\nkl}{\zl} \text{,}
\end{align}
where $\tildeCkl$ is the diagonal matrix defined analogously to $\Ckl$ in Equation \eqref{eq:Ckl} 
and $\hada$ is the Hadamard product. 
\subsection{Expected Log Likelihood Term \label{app:mc-grad-ell}}
Monte Carlo estimates of the gradients of the expected log likelihood term are:
\begin{align}
	\grad{\postmean{kj}} \elltermhat &= \frac{\pi_k}{\s}  \Kzzinv \sum_{n=1}^{\n}  \kzn \qfcovkjninv 
	\sum_{i=1}^{\s}  
	\left( \fnjki - \qfmeankjn \right) \log p(\yn | \fn^{(k,i)}) \text{,} \\
	\label{eq:gradcovell}
	 \grad{\postcov{kj}} \elltermhat &= \frac{\pi_k} {2 \s}   \sum_{n=1}^{\n}   \left( \ajn \ajn^T  \right) 
	 \sum_{i=1}^{\s}  \left[ \qfcovkjninvtwo \left( \fnjki - \qfmeankjn \right)^2 -  \qfcovkjninv \right] 
	 \log p(\yn | \fn^{(k,i)}) \text{,} \\
         \grad{\pi_k} \elltermhat  &=  \frac{1}{\s} \sum_{n=1}^\n \sum_{i=1}^{S} \log p(\yn | \fn^{(k,i)})		 
         \text{,}
\end{align}
and for diagonal covariance $\postcovdiag{kj}$ we replace $ \left( \ajn \ajn^T \right) $ 
	 { with } $\diag \left( \ajn \hada \ajn \right) $ in Equation \eqref{eq:gradcovell}, 
	 where $\hada$ is the Hadamard  product and $\diag(\vec{v})$ takes the input vector 
	 $\vec{v}$ and outputs a matrix with $\vec{v}$ on its diagonal.   We have also defined above 
	  $\kzn \defeq \kernel_j(\Z_j, \x_n)$, i.e.~the vector obtained from evaluating the covariance function $j$ 
	 between the inducing points $\Z_j$ and datapoint $\x_n$.  

\section{Gradients of the Evidence Lower Bound wrt Covariance Hyperparameters and Likelihood 
Parameters \label{app:grads-hyper-like}}
\newcommand{\gthetaj}{\grad_{\theta_j}}
Here we give the gradients of the variational objective ($\elbo$) wrt the covariance hyperparameters 
and, when required, the conditional likelihood parameters. We note here that our method does not 
require gradients of the conditional likelihood $\plike$ wrt the latent functions $\f$. However, if the conditional 
likelihood is parametrized by $\likeparam$ and point-estimates of these parameters are needed, these 
can also be learned in our variational framework.
\subsection{Covariance Hyperparameters \label{app:grads-hyper}} 
The gradients of the terms in the KL-divergence part of $\elbo$ wrt a covariance hyperparameter $\theta_j$ are:
\begin{align}
	\gthetaj \entermhat & = 0 \text{,}  \\
	\gthetaj  \crossterm &= - \frac{1}{2} \sum_{k=1}^{\k}  \pi_k \trace  
	\big[      \Kzzinv \gthetaj \Kzz  \\
	\nonumber
	&   \qquad  \qquad \qquad
	- \Kzzinv \gthetaj \Kzz \Kzzinv \left( \postmean{kj} \postmean{kj}^T + \postcov{j} \right) 
	\big] \text{.}
\end{align}
For the $\elltermhat$ we have that:
\begin{equation}
	\gthetaj \elltermhat = \sum_{n=1}^{\n} \sum_{k=1}^{\k} \pi_k \expectation{\qknf} { \gthetaj \log \qknf \log p(\yn | \fn) }  \text{,}
\end{equation}
and computing the corresponding gradient we obtain:
\begin{align}
\label{eq:gthetaell}
\gthetaj  \elltermhat = -\frac{1}{2} \sum_{n=1}^{\n} \sum_{k=1}^{\k} & \pi_k \expectationsingle{\qknf}
\Big[ \Big(   \qfcovkjninv \gthetaj \qfcovkjn \\
\nonumber
&  - 2 (f_{nj} - \qfmeankjn) \qfcovkjninv \gthetaj \qfmeankjn \\
\nonumber
	 & - (f_{nj} - \qfmeankjn)^2 \qfcovkjninvtwo \gthetaj \qfcovkjn \Big)  \log p(\yn | \fn)  \Big] \text{,}
\end{align}
for which we need:
%
%
\begin{align}
	\gthetaj \qfmeankjn &= \left( \gthetaj  \ajn^T \right)  \postmean{kj} \text{,} \\
	\gthetaj \qfcovkjn &=  \gthetaj  \matentry{ \priorcov }{n}{n} + 2 \left( \gthetaj \ajn^T \right) \postcov{kj} \ajn \\
	&= \gthetaj \knn - \left( \gthetaj \ajn^T \right) \kzn -  \ajn^T \gthetaj  \kzn	\\
	& \qquad + 2 \left( \gthetaj \ajn^T \right) \postcov{kj} \ajn \text{,}
\end{align}
where
\begin{equation}
	\label{eq:gradajn}
	\gthetaj  \ajn^T = \Big( \gthetaj \knz - \ajn^T  \gthetaj \Kzz \Big) \Kzzinv \text{,}
\end{equation}
and as in the main text we have defined $\ajn \defeq \matcol{\Aj}{n}$, i.e.~the vector corresponding
to the n$\mth$ column of $\Aj$.  Furthermore, as in the previous section, 
$\kzn \defeq \kernel_j(\Z_j, \x_n)$. 
\subsection{Likelihood Parameters \label{app:grads-like}}
Since the terms in the KL-divergence do not depend on the likelihood parameters 
we have that $\grad_{\likeparam}\entermhat = \grad_{\likeparam}\crossterm = 0$. For the gradients 
of the expected log likelihood term we have that:
\begin{equation}
\grad_{\likeparam}{\elltermhat} = 
	\frac{1}{\s} \sum_{n=1}^\n \sum_{k=1}^\k \pi_k \sum_{i=1}^{\s} \grad_{\likeparam} \log p(\yn | \fn^{(k,i)}, \likeparam) 
	\text{,}
\end{equation}
where $\{ \fn^{(k,i)} \} \sim \Normal(\fn; \qfmean{k(n)}, \qfcov{k(n)} )$,   for $k=1,\ldots, \k$ and
$i=1, \ldots, \s$.
\section{Gradients of the Log-Evidence Lower Bound wrt Inducing Inputs \label{app:grad-Z}}
These gradients can be obtained by using the same expression  as  the gradients
wrt covariance hyperparameters in \apref{app:grads-hyper}, considering the inducing inputs as additional hyperparameters
of the covariances. Therefore, we rewrite the gradients above keeping in mind that $\theta_j$ is an element 
of  inducing input  $\Z_j$, 
hence dropping those gradient terms that do not depend on $\Z_j$. As before we have that $\gthetaj \entermhat  = 0$.   
\begin{align}
\label{eq:grad-cross-hyper}
	\gthetaj  \crossterm = - \frac{1}{2} \sum_{k=1}^{\k}  \pi_k \trace  
	\big\{    
	\big[  \Kzzinv  
	-  \Kzzinv \left( \postmean{kj} \postmean{kj}^T + \postcov{j} \right) \Kzzinv 
	\big] \gthetaj \Kzz 
	\big\}\text{.}
\end{align}
Similarly, for the gradients of $\elltermhat$ in Equation \eqref{eq:gthetaell} we have that:
\begin{align}
	\gthetaj \qfmeankjn &= 
	\gthetaj \knz \Kzzinv  \postmean{kj}  -  \ajn^T  \gthetaj \Kzz  \Kzzinv  \postmean{kj}  \text{,}
\end{align}
and
\begin{align}
\gthetaj \qfcovkjn &= 
\gthetaj \knz \Big(- \Kzzinv   \kzn -  \ajn + 2  \Kzzinv  \postcov{kj} \ajn \Big) \\
& \qquad  + \ajn^T  \gthetaj \Kzz  \Kzzinv \kzn - 2  \ajn^T \gthetaj \Kzz \Kzzinv  \postcov{kj} \ajn
 \text{.}
\end{align}
\newcommand{\Ztilded}{\tilde{\Z}_{j}^{(d)}}
\newcommand{\Xtilded}{\tilde{\X}^{(d)}}
\newcommand{\vn}{\vec{v}_n}
\newcommand{\wn}{\vec{w}_n}
\newcommand{\V}{\mat{V}}
\newcommand{\termone}{t^{(1)}_n}
\newcommand{\vectermone}{\vec{t}^{(1)}}
\newcommand{\termtwo}{t^{(2)}_n}
\newcommand{\vectermtwo}{\vec{t}^{(2)}}

From the equations above, we see that in the computation of the gradients of $\elltermhat$ there are two types of terms. The first  
term is of the form:
\begin{align}
 \grad_{\theta} \termone \defeq 
 \vn^T \grad_{\theta} \Kzz \wn \text{,}
\end{align}
where we have dropped the index $j$ on the LHS of the equation for simplicity in the notation
and $\vn, \wn$ are $\m$-dimensional vectors.
%
Let $\V$ and $\W$ be the $\m \times \n$ matrices corresponding to the $\n$ vectors $\{ \vn \}$ and $\{ \wn \}$, respectively.
%
Furthermore, let us assume $\Ztilded$ is the $\m \times \m$ matrix of all pairwise differences on dimension $d$ of all inducing points
divided by the squared length-scale of the dimension  $\ell_d^2$, 
\begin{align}
\matentry{\Ztilded}{o}{p} = \frac{ \matentry{\Zj}{o}{d} - \matentry{\Zj}{p}{d}}{\ell_d^2} \text{.}
\end{align}
Hence, in the case of the squared exponential covariance function the above gradient can be calculated 
as follows (for all data points): 
\begin{align}
\grad_{\matcol{\Zj}{d}} \vectermone
=- ( (\Ztilded \hada \Kzz) \W ) \hada \V -  ( (\Ztilded \hada \Kzz) \V ) \hada \W \text{,}
\end{align}
where $\grad_{\matcol{\Zj}{d}}  \vectermone$ is the $\m \times \n$ matrix of gradients corresponding to dimension $d$ for all 
$m=1, \ldots, \m$ and $n=1, \ldots \n$.

Similarly, the second type of term is of the form:
\begin{align}
\grad_{\theta} \termtwo & \defeq \vn^T \grad_{\theta} \kzn  \text{,}
\end{align}
where $\vn$ and $\V$ are defined as before.  The gradients in the above expression 
wrt to the  inducing points (for all datapoints or a mini-batch) can be calculated as follows:
\begin{align}
	\grad_{\matcol{\Zj}{d}}{\vectermtwo} = - ( \Kxz  \hada \V^T \hada \Xtilded)^T \text{,}
\end{align}
where in the above equation $\Xtilded$
is the $\n \times \m$ matrix of all pairwise differences on dimension $d$ between all datapoints and 
 inducing points divided by the squared length-scale of the dimension  $\ell_d^2$:
\begin{align}
\matentry{\Xtilded}{o}{p} = \frac{\matentry{\X}{o}{d} - \matentry{\Zj}{p}{d}}{\ell_d^2} \text{.}
\end{align}
\section{Control Variates \label{app:ctrl-variates}}
We use control variates  \citep[see e.g.][\secrefout{8.2}]{ross2006simulation} 
to reduce the variance of the gradient estimates. In particular, we are interested in 
estimating  gradients of the form:
\begin{align}
\grad_{\lambda_k}  \expectation{\qknf}{\log p(\y_n | \fn)}
&=  \expectation{\qknf}{ g(\fn)  }\text{, with }  \\
 g(\fn) &=   \grad_{\lambda_k}  \log \qknf  \log p(\y_n | \fn)   \text{,} 
\end{align}
where the expectations are computed using samples from $\qknf$, which depends 
on the variational parameter $\lambda_k$. As suggested by \citet{ranganath2014black},
a sensible control variate is the so-called score function 
\begin{align}
h(\fn) =  \grad_{\lambda_k} \log \qknf \text{,}
\end{align}
whose expectation is zero.  Hence, the function:
\begin{equation}
\tilde{g}(\fn) = g(\fn)  - \hat{a} h(\fn) \text{,}
\end{equation}
has the same expectation as $g(\fn)$  but lower variance when $\hat{a}$ is given by:
\begin{align}
	\hat{a} = \frac{ \covariance[g(\fn), h(\fn)] } {\variance[ h(\fn) ]} \text{,}
\end{align} 
where $\covariance[g(\fn), h(\fn)]$ is the covariance between $g(\fn)$ and 
$h(\fn)$; $\variance[ h(\fn) ]$ is the variance  of $h(\fn)$; and 
both are estimated using samples from $\qknf$. Therefore, our corrected gradient 
is given by:
\begin{align}
\tilde{\grad}_{\lambda_k}  \expectation{\qknf}{\log p(\y_n | \fn)} & \defeq  \expectation{\qknf} { \tilde{g}(\fn)  } \\
& = \expectation{\qknf} {   \grad_{\lambda_k} \log \qknf ( \log p(\y_n | \fn)  - \hat{a}) } \text{.}
\end{align}
\section{Derivations for the Dense Posterior Case \label{app:nonsparse}}
In this section we consider the case of having a dense variational posterior, i.e.~not 
using the so-called sparse approximations.  We derive the expressions for the components of the log-evidence 
lower bound  and  show that the posterior parameters of the 
variational distribution $\qf$ are, in fact, of `free'-form.  Furthermore, we 
analyze the case of a Gaussian likelihood showing that, in the limit of a large number of samples, 
our estimates converge to the exact analytical solution.
\subsection{Evidence Lower Bound \label{app:nonsparse-elbo}}
Here we show the derivations of all the terms in $\elbo$ when considering 
the dense case, i.e.~$\m=\n$ and $\Z_j = \X$.  As described in the main text, 
the resulting changes to these terms can be obtained by replacing $\m$ with $\n$ and 
$\Kzz$ with $\Kxx$. Hence, for the terms in the KL-divergence part of $\elbo$ we have 
that:
\begin{align}
\label{eq:entropy-full}
\entermhat (\llambda)
&=
 - \sum_{k=1}^{\k} \pi_k \log \sum_{\ell=1}^\k \pi_{\ell} \Normal(\postmean{k}; \postmean{\ell}, \postcov{k} + \postcov{\ell}) 
	  \text{,} \\
\label{eq:cross-full}	  
	   \crossterm(\llambda) 
	   &= - \frac{1}{2} \sum_{k=1}^\k \pi_k \sum_{j=1}^\q [\n \log 2 \pi + \log \det{\Kxx} 
	+ \postmean{kj}^T \Kxxinv \postmean{kj} + \trace{\Kxxinv \postcov{kj}}], \quad 	
\end{align}
where we recall that now $\postmean{kj}$ and $\postcov{kj}$ are $\n$-dimensional objects. For 
the $\elltermhat$ we still need to compute empirical expectations over the 
variational distribution $\qf$, where the corresponding parameters are given by:
\begin{align}
\qfmean{kj} &= \Aj \postmean{kj} = \Kxx \Kxxinv \postmean{kj} = \postmean{kj} \text{,} \\
\qfcov{kj} &=  \priorcov - \Aj \Kxx + \Aj \postcov{kj} \Aj \\
&=  \Kxx - \Kxx \Kxxinv \Kxx +  \Kxx \Kxxinv \postcov{kj} \Kxx \Kxxinv \\
&= \postcov{kj} \text{.}
\end{align}
\subsection{Gaussian Likelihoods \label{app:gaussian-like}}
Consider the case of a single output and a single latent function ($\q=\p=1$) 
with a Gaussian conditional likelihood and a single full Gaussian variational 
posterior ($\k=1$).  
\begin{align}
	p(\y | \f) &= \Normal(\y; \f, \sigma^2 \I) \text{, and} \\
	\qf &= \Normal(\f; \postmean{}, \postcov{}) \text{.}
\end{align}
The entropy and the cross entropy terms are given in Equations 
\eqref{eq:entropy-full} and \eqref{eq:cross-full} with $\q = \k = 1$.
The expect log likelihood term can be determined analytically,
\begin{align}
\ellterm(\llambda) &= \sum_{n=1}^\n  \expectation{\qf}{ \log \pcond{y_n}{f_n, \likeparam} }  \text{,} \\
& = 
\log \Normal(\y; \postmean{}, \sigma^2 \I) - \frac{1}{2 \sigma^2} \trace\postcov{} \text{.}
\end{align}
Hence the gradients of $\elbo$ can also be determined analytically, yielding the 
optimal mean posterior  as follows:
\newcommand{\mopt}{\widehat{\postmean{}}}
\newcommand{\Sopt}{\widehat{\postcov{}}}
\begin{align}
\grad_{\postmean{}} \elbo & = \Kxxallinv \postmean{} + \frac{1}{\sigma^2} (\y - \postmean{}) = 0 \\
\left( \Kxxallinv + \frac{1}{\sigma^2} \right) \postmean{} &= \frac{1}{\sigma^2} \y \\
\mopt &= \left( \sigma^2 \Kxxallinv + \I \right)^{-1} \y \\
&= \Kxxall \left( \Kxxall + \sigma^2 \I \right)^{-1} \y \text{.}
\end{align}
Similarly for the posterior covariances we have that:
\begin{align}
\grad_{\postcov{}} \elbo &= \frac{1}{2} \postcov{}^{-1} - \frac{1}{2} \Kxxallinv - \frac{1}{2 \sigma^2} \I = 0 \\
\label{eq:optcov1}
\Sopt & = \left( \Kxxallinv + \frac{1}{\sigma^2} \I \right)^{-1} \\
\label{eq:optcov2}
&= \Kxxall - \Kxxall \frac{1}{\sigma^2} \left( \I + \Kxxall \frac{1}{\sigma^2} \right)^{-1} \Kxxall \\
&= \Kxxall - \Kxxall \left( \Kxxall + \sigma^2 \I \right)^{-1} \Kxxall \text{,}
\end{align}
where we have used Woodbury's formula to go from Equation \eqref{eq:optcov1} to
Equation \eqref{eq:optcov2}. The Equations above for the optimal posterior mean and 
posterior covariance, $\mopt$ and $\Sopt$, are the exact analytical expressions 
for regressions with Gaussian process priors and isotropic noise likelihoods, see
\citet[\secrefout 2.2]{rasmussen-williams-book}.

\section{Efficient Parametrization \label{app:efficient-param}}
In this appendix we prove Theorem \ref{th:efficient-param},  showing 
that it is possible to obtain  an efficient parameterization of the posterior covariances when 
using a full Gaussian approximation.  In this case we have that:
\begin{align}
\gsj \elbo & = \gsj \enterm + \gsj \crossterm + \gsj \ellterm \\
& = - \frac{1}{2} \Kzzinv + \frac{1}{2} \postcov{j}^{-1} + \sum_{n=1}^N \grad_{\postcov{j}}  \expectation{\qnf}{  \log \pcond{\yn}{\fn}}\\
\end{align}
Setting the gradients to zero  we have that
\begin{align}
  \frac{1}{2} \postcov{j}^{-1}  &=   \frac{1}{2}\Kzzinv    + \frac{1}{2} \Kzzinv \sum_{n=1}^{N} \kzn \lambdajn  \kznt \Kzzinv \text{,}
\end{align}
where 
\begin{align}
\lambdajn &= - {2} \fullderiv{\elln}{\Sigmann} \\
\elln &=  \expectation{\qnf}{  \log \pcond{\yn}{\fn}} \text{.}
\end{align}
 Therefore, the optimal solution for the posterior covariance is given by:
\begin{align}
\widehat{\postcov{j}} & =  \left(\Kzzinv +  \Kzzinv \Kzx \Lambdaj  \Kxz \Kzzinv \right)^{-1} \text{,} \\
\label{eq:Sopt}
&= \Kzz \left(\Kzz + \Kzx  \Lambdaj  \Kxz \right)^{-1} \Kzz \text{,}
\end{align}
where $\Lambdaj$ is a $\n \times \n$ diagonal matrix with  $\{\lambdajn\}_{n=1}^\n$ on the diagonal.
%
%
\QEDA
\section{Lower Variance of Mixture-of-Diagonals Posterior \label{app:modg-lower-variance}}
In this section we prove Theorem \ref{th:modg-lower-variance}.  
First we review Rao-Blackwellization \citep{casella1996rao}, which is also known as partial averaging or conditional Monte Carlo.
Suppose we want to estimate $V = \expectation{p(\x,\y)}{h(\X,\Y)}$ where $(\X, \Y)$ is a random variable with probability density $p(\x,\y)$ and $h(\X,\Y)$ is a random variable that is a function of $\X$ and $\Y$. 
It is easy to see that 
\begin{align}
\expectation{p(\x, \y)}{h(\X,\Y)}
=& \int p(\x, \y) h(\x, \y) \der \x \der \y \\
=& \int p(\y) p(\x | \y) h(\x, \y) \der \x \der \y \\
=& \expectation{p(\y)}{ \hat{h} (\Y) } \text{, with } \hat{h} (\Y) = \expectation{p(\x | \y)}{h(\X, \Y) | \Y}  \text{,}
\end{align}
and, from the conditional variance formula,
\begin{align}
\variance [\hat h(\Y)] < \variance [h(\X,\Y)].
\end{align}
Therefore when $\hat h(\Y)$ is easy to compute, it can be used to estimate $V$ with a lower variance than the original estimator.
When $p(\x,\y) = p(\x) p(\y)$, then $\hat h(\Y)$ is simplified to
\begin{align}
\hat h(\Y = \y) = \int p(\x) h(\x, \y) \der \x = \expectation{p(\x)}{h(\X, \Y) | \Y}.
\end{align}
\newcommand{\lamkn}{\llambda_{k(n)}}
\newcommand{\qknfull}{q_{k(n)} (\fn | \lamkn )}
\newcommand{\gradkn}{\grad_{\lamkn} \log \qknfull}
We apply Rao-Blackwellization to our problem with $\fn$ playing the role of the conditioning variable $\Y$ and 
$\fnotn$ playing the role of $\X$, where $\fnotn$ denotes $\f$ excluding $\fn$.
We note that this  Rao-Blackwellization analysis is only applicable to the dense case with  
a mixture of diagonals posterior, as it satisfies the independence condition, i.e.~$p(\x,\y) = p(\x) p(\y)$.

First, we express the gradient of $\lamkn$ as an expectation by interchanging the integral and gradient operators giving
\begin{align}
\grad_{\lamkn} \expectation{\qkf} {\log \plike}
=& \expectation{\qkf} { \grad_{\lamkn} \log \qkf  \log \plike }.
\end{align}
The Rao-Blackwellized estimator is thus
\begin{align}
\hat h(\fn) 
& = \int q(\fnotn) \grad_{\lamkn} \log \qkf \log p(\y | \f) \der \fnotn \\
&= \int q(\fnotn) \gradkn  \log p(\y | \f) \der \fnotn \\
\label{eq:iid-like-th3}
&= \gradkn  \int q(\fnotn) \left[  \log p(\yn | \fn)  + \log p(\ynotn | \fnotn) \right]  \der \fnotn   \\
&= \gradkn  \left[ \log p(\yn | \fn) + C \right],
\end{align}
where we have used the factorization of the conditional likelihood in 
Equation \eqref{eq:iid-like-th3},  and where  $C$ is a constant w.r.t $\fn$. 
This gives the Rao-Blackwellized gradient,
\begin{align}
\grad_{\lamkn} \expectation{\qkf}{ \log p(\y | \f )} 
&= \expectation{\qknfull}{\hat h(\fn)} \\
&= \expectation{\qknfull}{\gradkn \log p(\y_n | \fn)} \text{,}
\end{align}
where we have used the fact that $\expectation{q}{\grad \log{q}} = 0$ for any $q$.
We see then that the expression above is exactly the gradient  obtained in Equation \eqref{eq:grad-full-2}. 
\QEDA

\section{Details of Complexity analysis \label{app:complexity}}
Here we give the details of the computational complexity of a single evaluation 
of the $\elbo$ and its gradients. 
Let $\A$ be some matrix of size $p \times q$, $\B$
be some matrix of size $q \times r$, $\C$ be some matrix of size $q \times p$, and $\D$ be a square matrix
of size $s \times s$. Then we assume that $T(\A\B) \in \bigO(pqr)$, $T(\D^{-1}) \in \bigO(s^3)$, $T(\det{\D})
\in \bigO(s^3)$, $T(\diag(\A\C)) \in \bigO(pq)$, and $T(\trace(\A\C)) \in \bigO(pq)$.

%

\subsection{Interim values}
We first provide an analysis of the cost of various interim expressions that get re-used throughout the
computation of the $\elbo$. The interim expressions include: various values computed from the kernel functions
($\Kzz$, $\Kzzinv$, $\det{\Kzz}$, $\kzn$), values used to determine the shape of the Gaussian distributions from which we
sample ($\qfmeankjn$, $\qfcovkjn$), and various intermediate values ($\matentry{ \priorcov }{n}{n}$, $\ajn$).
We note that expressions with a subscript or superscript of $j$, $n$, or $k$ are dependent on $\q$, $\bsize$, and $\k$
respectively. For example, in the case of $\kzn$, we actually need to consider the cost of $\q \times \bsize$ vectors
of size $\m$. 

As stated in the main text, we assume that the kernel function is simple enough such that evaluating its output between two points is in $\bigO(\d)$. Hence the cost of evaluating interim values is given by:
\begin{align}
\nonumber
    T(\Kzz) &\in \bigO(\q\m^2 \d) \text{,}\\ \nonumber
    T(\Kzzinv) &\in \bigO(\q\m^3) \text{,}\\ \nonumber
    T(\det{\Kzz}) &\in \bigO(\q\m^3) \text{,}\\ \nonumber
    T(\kzn) &\in \bigO(\q\bsize\m\d) \text{,}\\ \nonumber
    T(\qfmeankjn) &\in \bigO(\k\q\bsize\m) \text{,}\\ \nonumber
    T(\qfcovkjn) &\in \bigO(\k\q\bsize\m^2) \text{,}\\ \nonumber
    T(\matentry{ \priorcov }{n}{n}) &\in \bigO(\q\bsize\m) \text{,}\\ \nonumber
    T(\ajn) &\in \bigO(\q\bsize\m^2)\text{,}\\  \nonumber
  	\nonumber & \text{ and for diagonal covariance we have:} \\ \nonumber
    T(\qfcovkjn) &\in \bigO(\k\q\bsize\m) \text{.}\\ \nonumber
\end{align}
The computational cost of evaluating all interim values is thus in $\bigO(\q\m(\m\d + \m^2 + \bsize\d + \k\bsize\m))$ for
the case of full covariance and in $\bigO(\q\m(\m\d + \m^2, \bsize\d, \k\bsize))$ for diagonal covariances.

\subsection{Analysis of the $\klterm$}
The computational complexity of terms in the KL-divergence and its gradients is given by:
\begin{align}
 \nonumber
    T(\crossterm) &\in \bigO(\k\q\m^2) \text{,}\\ \nonumber
    T(\enterm) &\in \bigO(\k^2\q\m^3) \text{,}\\ \nonumber
    T(\grad_{\postmean{}}\crossterm) &\in \bigO(\k\q\m^2) \text{,}\\ \nonumber
    T(\grad_{\postcov{}}\crossterm) &\in \bigO(\k\q\m^2) \text{,}\\ \nonumber
    T(\grad_{\pi_k}\crossterm) &\in \bigO(\k\q\m^2) \text{,}\\ \nonumber
    T(\grad_{\postmean{}}\enterm) &\in \bigO(\k^2\q\m^3) \text{,}\\ \nonumber
    T(\grad_{\postcov{}}\enterm) &\in \bigO(\k^2\q\m^3) \text{,}\\ \nonumber
    T(\grad_{\pi_k}\enterm) &\in \bigO(\k^2\q\m^3) \text{,}\\ \nonumber
  	\nonumber & \text{ and for diagonal covariance we have:} \\ \nonumber
    T(\grad_{\postcovdiag{}}\crossterm) &\in \bigO(\k\q\m) \text{,}\\ \nonumber
    T(\grad_{\postmean{}}\enterm) &\in \bigO(\k^2\q\m) \text{,}\\ \nonumber
    T(\grad_{\postcovdiag{}}\enterm) &\in \bigO(\k^2\q\m) \text{,}\\ \nonumber
    T(\grad_{\pi_k}\enterm) &\in \bigO(\k^2\q\m) \text{.}
\end{align}
Hence the computational cost of evaluating the KL-term and its gradients is in $\bigO(\k^2\q\m^3)$ for full
covariances, and $\bigO(\k\q\m(\k + \m))$ for diagonal covariances.

\subsection{Analysis of the $\ellterm$}
Let $T(\logpliken) \in \bigO(\likecost)$, then the cost of evaluating the expected log likelihood and its gradients
is given by:
\begin{align}
 \nonumber
    T(\ellterm) &\in \bigO(\k\bsize\s(\likecost + \q)) \text{,}\\ \nonumber
    T(\grad_{\postmean{}}\ellterm) &\in \bigO(\k\q(\m^2 + \bsize\m + \bsize\s\likecost)) \text{,}\\ \nonumber
    T(\grad_{\postcov{}}\ellterm) &\in \bigO(\k\q\bsize(\m^2 + \s\likecost)) \text{,}\\ \nonumber
    T(\grad_{\pi_k}\ellterm) &\in \bigO(\k\bsize\s(\likecost + \q)) \text{,}\\ \nonumber
  	\nonumber & \text{ and for diagonal covariance we have:} \\ \nonumber
    T(\grad_{\postcovdiag{}}\ellterm) &\in \bigO(\k\q\bsize(\m + \s\likecost)) \text{.}
\end{align}
Hence the computational cost of evaluating the ell term and its gradients is in $\bigO(\k\q\bsize(\m^2 + \s\likecost))$
for full covariances, and $\bigO(\k\q\bsize(\m + \s\likecost))$ for diagonal covariances.

%
%
\section{Additional Results}
\subsection{Small-scale Experiments --- Warped GPs} \label{sec:smal-expts-creep}
\begin{figure}[h!]
	\centering
	\begin{tabular}{c}
		\includegraphics[scale=1]{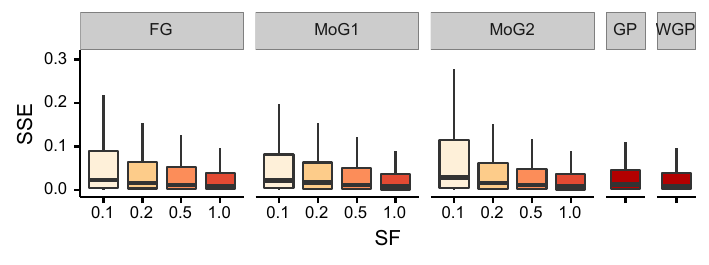} \\
		\includegraphics[scale=1]{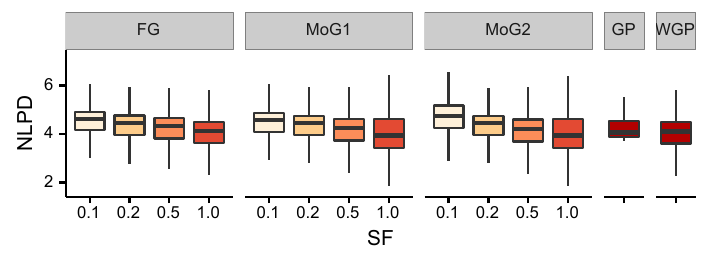}     
	\end{tabular}
	\caption{%
		The distributions of \sse and \nlpd for a warped Gaussian process likelihood model on the \creep dataset.
		Three approximate posteriors in \savigp are used:
		\full (full Gaussian),
		\mix (diagonal Gaussian), and 
		\mixtwo (mixture of two diagonal Gaussians), along with
		various sparsity factors ($\sfmath = M/N$). 
		The smaller the \sftext the sparser the model, with $\sfmath=1$ corresponding 
		to the original model without sparsity. \wgp corresponds to the performance of the exact inference method for warped Gaussian process models \citep{snelson2003warped}, and \gptext is the performance of the exact inference on a univariate Gaussian likelihood model.
	}
	\label{fig:creep}%
\end{figure}

In the case of the \creep dataset (Figure~\ref{fig:creep}), \sse and \nlpd are better in denser models. As expected, the performance of the dense ($\sfmath=1$) \full model  is identical to the performance of \wgp. We also note that \nlpd in \gptext has less variation compared to the other models, which can be attributed to its assumption of Gaussian noise.

\subsection{Medium-scale Experiments --- Binary Classification}\label{sec:medium-binary-mnistbin}
\begin{figure}[h!]
	\centering
	\begin{tabular}{cc}
		\includegraphics[scale=1]{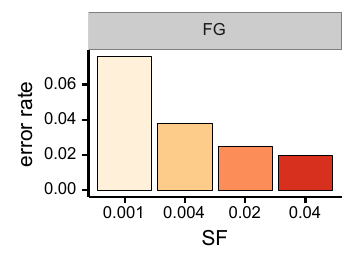} &
		\includegraphics[scale=1]{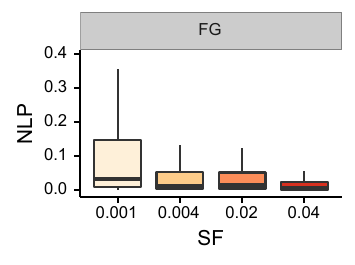} 		
	\end{tabular}
	\caption{Error rates and 
		\nlp for binary classification on the 
		\mnistbin dataset. 
		We used a full Gaussian (\full) posterior approximation across
		various sparsity factors ($\sfmath = M/N$). 
		The smaller the SF the sparser the model.
	}
	\label{fig:mnist2_bin}%
\end{figure}
We see in Figure~\ref{fig:mnist2_bin} that the performance improves significantly with denser models, i.e.~when 
we use a larger number of inducing variables. 
Overall, our model achieves an accuracy of $98.1\%$ and a mean \nlp of $0.062$ on \mnistbin
with a sparsity factor of $0.04$. This is slightly better than the performance 
obtained by \citet{hensman-et-al-aistats-2015}, who report an accuracy of $97.8\%$ and \nlp of $0.069$, although they used a smaller number of inducing variables. 
As we have seen in \secref{sec:expts-learn-z}, our model also achieves similar performance when using the same sparsity factor 
as that  used by \citet{hensman-et-al-aistats-2015} when the inducing inputs 
(i.e.~the locations of the inducing variables) are learned.

\subsection{Medium-scale Experiments --- Gaussian Process Regression Networks}\label{sec:expts-medium-gprn-sarcos}
\begin{figure}[t]
	\centering
	\begin{tabular}{cc}
		\includegraphics[]{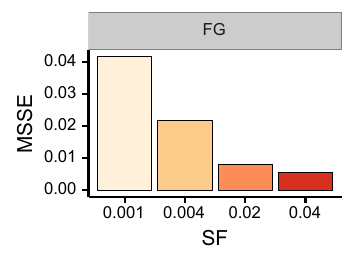} &
		\includegraphics[]{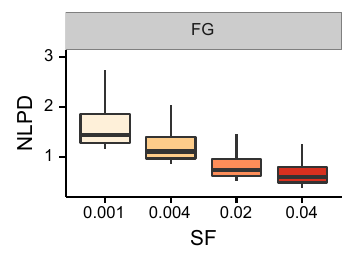} 		
	\end{tabular}
	\caption{Mean \sse and 
		\nlpd for multi-output regression on the 
		\sarcos dataset. 
		We used a full Gaussian (\full) posterior approximation across
		various sparsity factors ($\sfmath = M/N$). 
		The smaller the SF the sparser the model.
	}
	\label{fig:sarcos_all}%
\end{figure}
Figure~\ref{fig:sarcos_all} shows the performance on the \sarcos dataset.
Compared to \sarcostwo, learning all 
joints yields a slight increase in \msse for joints $4$ and $7$, but also resulted in a significant decrease in \nlpd. 
This shows that the
transition from learning only $2$ joints to all $7$ joints (as in \sarcostwo), while unhelpful for point estimates, led to a significantly better posterior density estimate. 
We also note that we had to make use of $10,000$ samples for this dataset
to get  
stable results over different sparsity factors. 
This is a much larger number than the $100$  samples used in \mnist. 
This is likely because the \gprn{} likelihood model involves multiplication between latent processes, 
which increases the variance of sampling results by a squared factor, and therefore, it is reasonable to 
require more samples in this complex likelihood model.

\subsection{Medium-scale Experiments --- Inducing-input Learning}\label{sec:expts-medium-inducing-mnist-sarcos}
\begin{figure}[h!]
	\centering
	\begin{tabular}{cc}
		\includegraphics[]{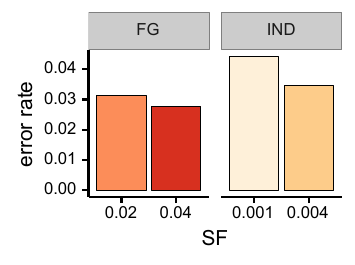} &
		\includegraphics[]{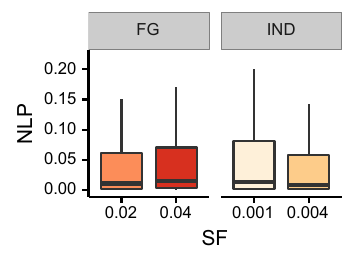} 		
	\end{tabular}
	\caption{Comparison of error rate and 
		\nlp obtained by \savigp without learning (\full) and with learning (\ind)
		of inducing inputs for multi-class classification on the 
		\mnist dataset.}
	\label{fig:mnist_ind}%
\end{figure}
\begin{figure}[h!]
	\centering
	\begin{tabular}{cc}
		\includegraphics[]{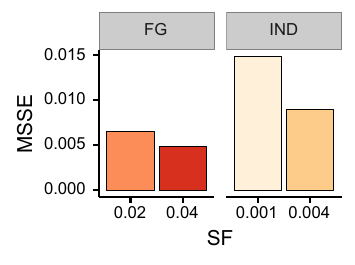} &
		\includegraphics[]{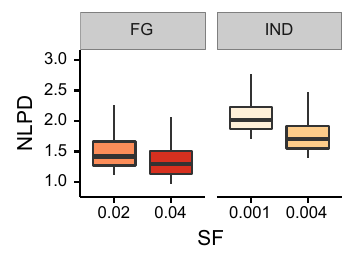} 		
	\end{tabular}
	\caption{Comparison of mean \sse and 
		\nlpd  obtained by \savigp without learning (\full) and with learning (\ind)
		of inducing points for multi-output regression on the 
		\sarcostwo dataset.}
	\label{fig:sacros2_ind}%
\end{figure}

\begin{figure}[h!]
	\centering
	\begin{tabular}{cc}
		\includegraphics[]{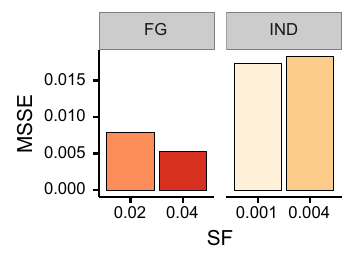} &
		\includegraphics[]{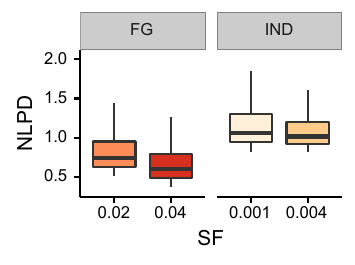} 		
	\end{tabular}
	\caption{Comparison of mean \sse and 
		\nlpd  obtained by \savigp without learning (\full) and with learning (\ind)
		of inducing inputs for multi-output regression on the 
		\sarcos dataset.}
	\label{fig:sacros_ind}%
\end{figure}
Figure~\ref{fig:mnist_ind} shows the performance of the model under the two settings (with and without learning inducing inputs) on \mnist. As with \mnistbin, we see that learning the location of the inducing variables yields a large gain in performance. In fact, the sparser models with inducing point learning performed similarly to the denser models, despite the fact that the two models differed by an order of magnitude when it came to the number of inducing variables. Figures~\ref{fig:sacros2_ind}, \ref{fig:sacros_ind} show the performance on \sarcos and \sarcostwo. We observe an improvement in performance from learning inducing inputs. However, the improvement is less significant than for \mnist, which shows that different datasets have significantly different sensitivity to the location of the inducing variables.

\section{Notation \label{app:notation}}
A summary of the notation used in this paper is shown in Table \ref{tab:notation}.
\begin{table}[h!]
\begin{center}
\caption{Summary of notation used in the paper.}
\label{tab:notation}
\begin{tabular}{l l }
\toprule
Symbol  & Description \\
\midrule
$\vec{v}$ & Bold lower case denotes a vector \\
$\C$ & Bold upper case denotes a matrix \\ 
$\det{\C}$ & Determinant of matrix $\C$ \\
$\trace{\C}$ & Trace of matrix $\C$ \\
$\d$ & Input dimensionality \\
$\n$ & Number of training datapoints \\
$\m$ & Number of inducing variables per latent process \\
$\p$ & Number of outputs \\
$\q$ & Number of latent functions \\
$\k$  & Number of mixture components in variational posterior \\
$\s$ & Number of samples in Monte Carlo estimates \\
$\x_n$ & n$\mth$ input datapoint \\
$\X$ & $\n \times \d$ matrix of input data \\
$\yn$ & $\p$-dimensional vector of outputs (label) for n$\mth$ observation\\
$\fn$ & $\q$-dimensional vector of latent function values for n$\mth$ observation \\
$\fj$ & $\n$-dimensional vector of latent function values for $j\mth$ process \\
$\uj$ & Inducing variables of latent process $j$ \\
$\u$ & Vector of all inducing variables \\
$\Zj$ & $\m \times \d$ Matrix of inducing inputs for latent process $j$ \\
$\kernel_j(\x, \x^\prime)$ & Covariance function of latent process $j$ evaluated at $\x, \x^\prime$ \\
$\Kxx$ & $\n \times \n$ covariance matrix obtained by evaluating $\kernel_j(\X, \X)$ \\
$\Kzz$ &  $\m \times \m$ covariance matrix obtained by evaluating $\kernel_j(\Zj, \Zj)$\\
$\Kzzall$ & Block-diagonal covariance with blocks $\Kzz$ \\  
$\Kxz$ & $\n \times \m$ covariance between $\X$ and $\Zj$ \\
$\kzn$ & $\m \times 1$ covariance vector between $\Zj$ and $\x_n$ \\
$\qu$ & Variational distribution over inducing variables \\
$\qf$ & Variational distribution over latent functions \\
$\postmean{}$ & Posterior mean over inducing variables \\
$\postcov{}$ & Posterior covariance over inducing variables \\
$\qfmean{}$ & Posterior mean over latent functions \\
$\qfcov{}$ & Posterior covariance over latent functions \\
\bottomrule
\end{tabular}
\end{center}
\end{table}

\clearpage
\bibliography{refs-savgip-jmlr}

\end{document}